%% file: main.tex
\pdfoutput=1
\documentclass[11pt]{article}
\usepackage{acl}
\usepackage{times}
\usepackage{latexsym}
\usepackage[T1]{fontenc}
\usepackage[utf8]{inputenc}
\usepackage{microtype}
\usepackage{inconsolata}
\usepackage{graphicx}

\usepackage{ulem}  

\usepackage{amsmath}
\usepackage{makecell}  

\usepackage{booktabs}
\usepackage{amssymb}  
\usepackage{multirow} 
\usepackage{lscape}   
\usepackage{array}
\usepackage[table,xcdraw]{xcolor} 
\usepackage{cite}
\usepackage{tcolorbox}

\usepackage{listings}

\definecolor{lightgray}{gray}{0.95}

\usepackage{minitoc}
\usepackage{stackengine}

\newcommand{\betterdoubleuline}[1]{%
  \stackengine{-1.5pt}{\uline{\textbf{\textit{#1}}}}{\smash{\rule{\widthof{\textbf{\textit{#1}}}}{0.4pt}}}{O}{c}{F}{F}{L}%
}
\newcommand\footnoteONLYtext[1]{
    \let \mybackup \thefootnote
    \let \thefootnote \relax
    \footnotetext{#1}
    \let \thefootnote \mybackup
    \let \mybackup \imareallyundefinedcommand}

\lstdefinelanguage{json}{
    morestring=[b]",
    morecomment=[l]{//},
    commentstyle=\color{gray},
    stringstyle=\color{blue},
    literate=
     *{0}{{{\color{black}0}}}{1}
      {1}{{{\color{black}1}}}{1}
      {2}{{{\color{black}2}}}{1}
      {3}{{{\color{black}3}}}{1}
      {4}{{{\color{black}4}}}{1}
      {5}{{{\color{black}5}}}{1}
      {6}{{{\color{black}6}}}{1}
      {7}{{{\color{black}7}}}{1}
      {8}{{{\color{black}8}}}{1}
      {9}{{{\color{black}9}}}{1}
}

\newtcolorbox[auto counter, number within=section, list type=subsubsection, list inside=toc]{sectionbox}[2][]{
  colback=white!98!gray, 
  colframe=black, 
  colbacktitle=white!90!gray, 
  coltitle=black, 
  fonttitle=\bfseries,
  title={#2}, 
  list entry={Comment \thetcbcounter\quad}
}

\title{Online-PVLM: Advancing Personalized VLMs with Online Concept Learning}

\author{
\textbf{Huiyu Bai}{\footnotesize$^{1\dagger{\sharp} }$} \quad 
\textbf{Runze Wang}{\footnotesize$^{2\dagger}$} \quad 
\textbf{Zhuoyun Du}{\footnotesize$^{3\dagger{\sharp}}$} \quad 
\textbf{Yiyang Zhao}{\footnotesize$^{1}$} \\
\textbf{Fengji Zhang}{\footnotesize$^{4}$} \quad 
\textbf{Haoyu Chen}{\footnotesize$^{5}$} \quad 
\textbf{Xiaoyong Zhu}{\footnotesize$^{2}$} \quad 
\textbf{Bo Zheng}{\footnotesize$^{2}$} \quad 
\textbf{Xuejiao Zhao}{\footnotesize$^{1\text{*}}$} \\[3pt]
$^{1}$Nanyang Technological University \quad 
$^{2}$Alibaba Group \quad 
$^{3}$Zhejiang University \\
$^{4}$City University of Hong Kong \quad 
$^{5}$University of Oulu \\[3pt]
\texttt{huiyu001@e.ntu.edu.sg, xjzhao@ntu.edu.sg}
}

\begin{document}

\maketitle



\footnoteONLYtext{$^\dagger$Equal Contribution.}
\footnoteONLYtext{$^{\sharp}$Work done during an internship at Alibaba Group.}
\footnoteONLYtext{$^*$Corresponding Authors: Xuejiao Zhao.}

\newcommand{\encoder}{Omni Concept Embedder}
\input{sec/0_abstract}
\input{sec/1_introduction}

\input{sec/2_relatedwork}

\input{sec/3_method}

\input{sec/4_dataset}

\input{sec/5_experiment}

\input{sec/6_ablation}

\input{sec/7_conclusion}

\input{sec/8_limitation}

\bibliography{custom}
\clearpage
\appendix
\input{sec/9_appendix}

\end{document}

%% file: sec/0_abstract.tex
\begin{abstract}
Personalized Visual Language Models (VLMs) are gaining increasing attention for their formidable ability in user-specific concepts aligned interactions (e.g., identifying a user's bike). Existing methods typically require the learning of separate embeddings for each new concept, which fails to support real-time adaptation during testing. This limitation becomes particularly pronounced in large-scale scenarios, where efficient retrieval of concept embeddings is not achievable. To alleviate this gap, we propose Online-PVLM, a framework for online concept learning by leveraging hyperbolic representations. Our approach makes a train-free paradigm for concept embeddings generation at test time, making the use of personalized VLMs both scalable and efficient.
In addition, we develop OP-Eval, a comprehensive and large-scale benchmark comprising 1,292 concepts and over 30K high-quality instances with diverse question types, designed to rigorously assess online concept learning in realistic scenarios. Extensive experiments demonstrate the state-of-the-art performance of our proposed framework. Our source code and dataset will be made available.
\end{abstract}
\begin{figure}[h!]  
    \includegraphics[width=\linewidth]{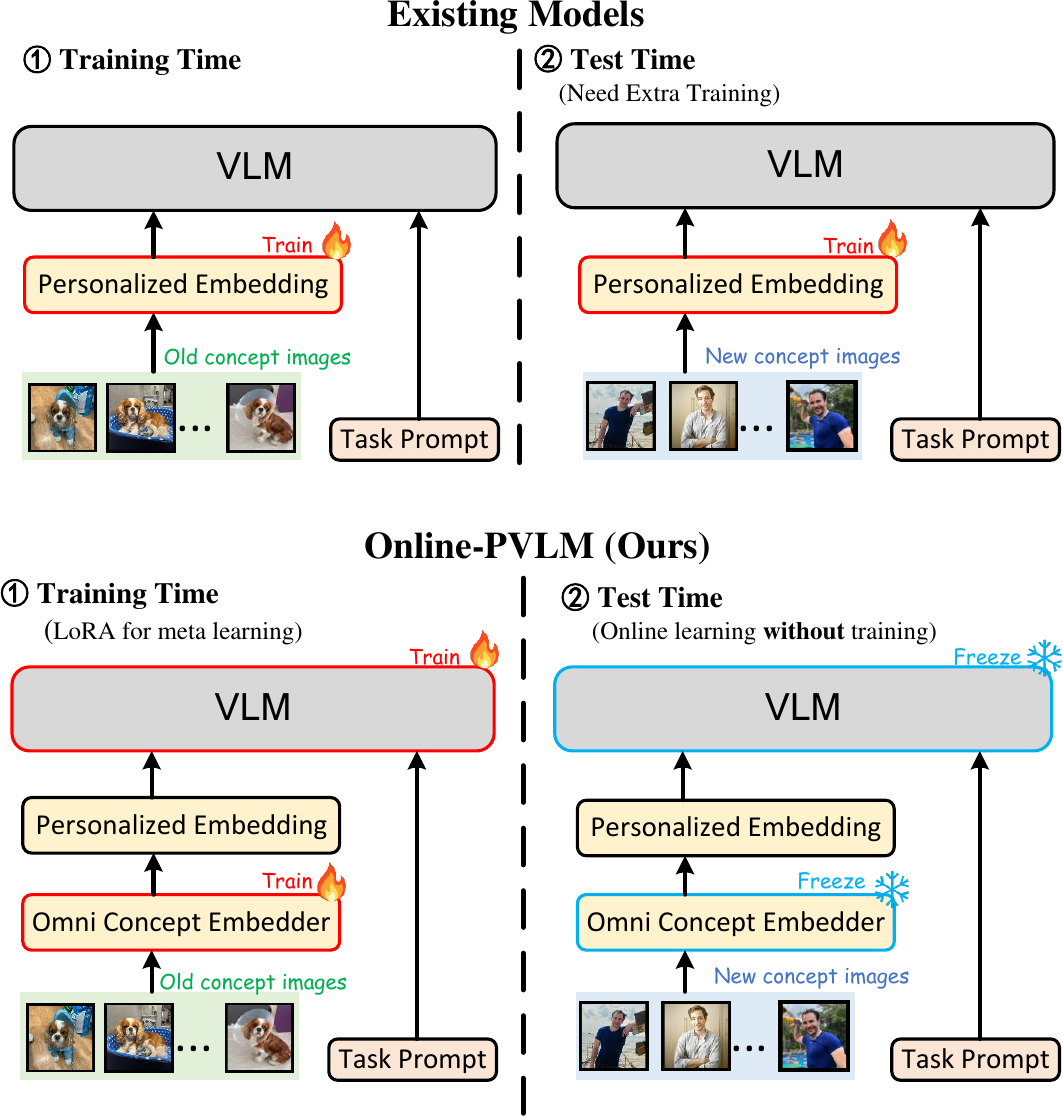}  
    \caption{An illustrative comparison of our work with existing concept learning methods, e.g., \citet{nguyen2024yo} and \citet{alaluf2024myvlm}. Our proposed Online-PVLM is capable of generating personalized concept embedding for new concepts in an online learning manner without further training at test time.}
    \label{fig:first_look}
\vspace{-10pt}
\end{figure}

%% file: sec/1_introduction.tex
\section{Introduction}

Vision-language models (VLMs) have demonstrated formidable generic capabilities \citep{achiam2023gpt, chen2024internvl, liu2023visual, liu2024deepseek}, but they are inadequate for handling personalized queries.  For example, when shown an image of <my\_brother>\footnote{\texttt{<my\_brother>} is a user-defined identifier representing the user's brother.} celebrating his graduation, a typical VLM might describe the scene as “a man in a cap and gown”, whereas a personalized VLM should recognize the specific context and respond with "<my\_brother> is finally graduating today." In this context, <my\_brother> represents a user-specific \betterdoubleuline{\textit{\textbf{concept}}}. Recognizing and interpreting such customized concepts is increasingly vital as personalized applications in personal assistance, and content generation grow rapidly \citep{li2023llava, li2024truthreader, yunusov2024mirrorstories}.

Several efforts have been made in personalized VLMs: \citet{nguyen2024yo} trains concept-specific embeddings and uses them as soft prompts, \citet{alaluf2024myvlm} concatenates concept embeddings with image features, and \citet{pi2024personalized} explores visual instruction tuning to learn personalized information online. Despite these works showing promising progress in building personalized VLMs \citep{alaluf2024myvlm, nguyen2024yo}, these approaches are inherently constrained with several limitations, hindering real-world deployment: \textbf{1. Lack of Memory:} Techniques like instruction tuning \citep{pi2024personalized} fail to store or recall information from previous interactions with individual users, preventing long-term personalization; \textbf{2. Inability to Learn Online:} Many methods \citep{alaluf2024myvlm, nguyen2024yo, an2024mc} that generate individual-specific embeddings are limited by static knowledge, requiring test-time training for new concept adaptation; \textbf{3. Scalability Issues:} Concept-specific training approaches \citep{nguyen2024yo, alaluf2024myvlm} often rely on numerous negative samples, leading to high computational costs and limited scalability for large-scale personalization. Addressing these challenges is essential for advancing practical personalized multimodal systems \citep{qian2024scaling}.

Thus, we argue that an effective personalized VLM for practical use should exhibit the following key traits:
\textbf{1. Dynamic memory retention:} The ability to store and recall user-specific features from prior interactions, enabling adaptation to evolving contexts;
\textbf{2. Online learning capability:} Seamless integration of new concepts without requiring re-training for each instance;
\textbf{3. Scalable personalization:} Efficient adaptation across diverse users with minimal computational overhead.

\input{tables/method_comparison}

Meeting these criteria requires a paradigm transformation from static, training-heavy methods to an adaptive, memory-augmented framework. Accordingly, we introduce Online-PVLM (Figure \ref{fig:first_look}), a personalized VLM framework designed for online concept learning. Motivated by both our preliminary experiments and the findings of \citet{pi2024personalized}, which demonstrate the effectiveness of Low-Rank Adaptation (LoRA) \citep{hu2022lora} as an online learner, we adopt LoRA as the foundation of our framework. On this basis, Online-PVLM integrates three core components: a lightweight \encoder{} for generating concept-specific embeddings on the fly, a hyperbolic discriminator for improved representation learning via concept–image matching, and a LoRA-enhanced VLM for instruction following. As shown in Table \ref{tab:methods_comparison}, Online-PVLM’s modular design enables efficient personalized understanding with minimal training data across varied scenarios, outperforming existing methods that rely on extensive fine-tuning or large-scale personalized datasets.

To assess online concept learning in realistic settings, we introduce OP-Eval—a benchmark designed for high-throughput concepts, cross-concept queries, and diverse downstream tasks. It includes 1,292 concepts and ~30,000 annotations, reflecting real-world scale and sparsity. OP-Eval spans single- and multi-concept protocols across varied question types, and supports three representative tasks: QA, identification, and captioning. Together, these elements make OP-Eval a faithful proxy for real-world deployment.

In summary, our contributions are threefold:
\begin{enumerate}
\item  We propose the task of online concept learning, which is the first to move the field beyond per‑concept embedding training toward large‑scale generalization. 

\item  We introduce OP-Eval, a high-quality benchmark that realistically simulates online concept learning scenarios through large-scale, sparse, and diverse concept-task configurations.


\item We propose Online-PVLM —an innovative VLM architecture featuring hyperbolic representations for effective personalization that can handle a consistently increasing number of concepts. Extensive experimental results show our method achieves state-of-the-art performances over various benchmarks.

\end{enumerate}

%% file: tables/method_comparison.tex
\begin{table}[t!]
\resizebox{\linewidth}{!}{
\begin{tabular}{lcccc}
\hline
\textbf{Method} & \makecell{\textbf{Online}\\\textbf{Learning}} & \makecell{\textbf{Concept}\\\textbf{Rep.}} & \textbf{Multi.} & \makecell{\textbf{Imgs per}\\\textbf{Con.}} \\
\hline
MyVLM \citep{alaluf2024myvlm}& \includegraphics[height=8pt]{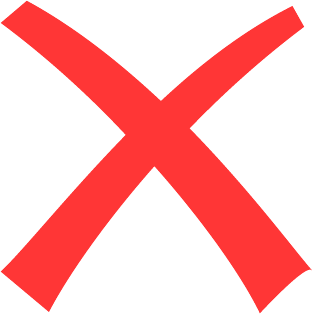} & \includegraphics[height=8pt]{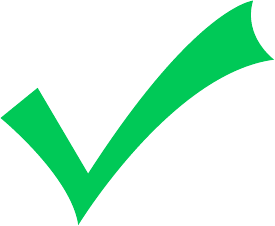} & \includegraphics[height=8pt]{images/false.pdf} & $\sim10^2$ \\
Yo'LLaVA \citep{nguyen2024yo} & \includegraphics[height=8pt]{images/false.pdf} & \includegraphics[height=8pt]{images/correct.pdf} & \includegraphics[height=8pt]{images/correct.pdf} &  $\sim10^2$ \\
MC-LLaVA \citep{an2024mc} & \includegraphics[height=8pt]{images/false.pdf} & \includegraphics[height=8pt]{images/correct.pdf} & \includegraphics[height=8pt]{images/correct.pdf} & $\sim10^1$ \\
P-LLaVA \citep{pi2024personalized} & \includegraphics[height=8pt]{images/correct.pdf} & \includegraphics[height=8pt]{images/false.pdf} & \includegraphics[height=8pt]{images/correct.pdf} & $\sim10^1$ \\
Online-PVLM (Ours) & \includegraphics[height=8pt]{images/correct.pdf} & \includegraphics[height=8pt]{images/correct.pdf} & \includegraphics[height=8pt]{images/correct.pdf} & $\sim10^1$ \\
\hline
\end{tabular}
}
\caption{\small \textbf{Comparison of previous methods with Online-PVLM.} \textbf{Concept Rep.} indicates use of concept embeddings; \textbf{Multi.} denotes support for multi-concept tasks; \textbf{Imgs per Con.} includes images needed per concept for training (both positive images and negative images). Our approach demonstrates enhanced performance across all key aspects compared to existing methods.}
\label{tab:methods_comparison}
\vspace{-10pt}
\end{table}

%% file: sec/2_relatedwork.tex
\section{Related Work}
\noindent\textbf{Personalizing VLM.}
Visual language models (VLMs) have progressed from early efforts in modality alignment for tasks like VQA and image–text retrieval \citep{hudson2019gqa, plummer2015flickr30k} to more recent instruction-following systems with enhanced visual reasoning \citep{hong2024cogagent, yang2025r1}. Despite these advances, personalized visual understanding remains underexplored. Recent approaches such as MyVLM \citep{alaluf2024myvlm} and Yo’LLAVA \citep{nguyen2024yo} inject user-specific concepts via tunable modules or additional tokens. Other methods, like PeKit \citep{hao2024rememberretrievegenerateunderstanding} and MC-LLAVA \citep{an2024mc}, explore retrieval-augmented generation and multi-concept modeling. However, most of these approaches rely on per-user training, limiting scalability. PVIT \citep{pi2024personalized} proposes an online learning dataset (P-Bench) to alleviate this but it suffers from noisy supervision and limited model expressiveness. Moreover, the field lacks a standardized, task-diverse benchmark for evaluating personalized VLMs at scale.
\vspace{1em}

\noindent\textbf{Online Representation Learning for VLM.}
Online representation learning for VLMs has advanced through prompt tuning and adapter-based methods. Prompt-based approaches, such as CoOp \citep{zhou2022cocoop} and ProDA \citep{zhang2021prototypical}, replace static templates with learnable vectors, improving flexibility at the cost of generalization. Extensions like CoCoOp and MetaPrompt \citep{Wang2024LearningTL} incorporate visual or meta-learning signals to enhance domain adaptability. Adapter-style methods, including CLIP-Adapter \citep{gao2021clip} and Tip-Adapter \citep{zhang2021tip}, insert lightweight modules into frozen backbones for efficient fine-tuning, while MMA \citep{song2019combining} aligns multimodal features to improve coherence. Despite these advances, personalized representation learning remains underexplored. Although Yo’LLAVA \citep{nguyen2024yo} supports concept embedding generation, it struggles with handling unseen concepts and lacks general-purpose embedding capabilities.

%% file: sec/3_method.tex
\section{Methodology}
\subsection{Problem Formulation}
\begin{figure*}[t]  
    \centering
    \includegraphics[width=\textwidth]{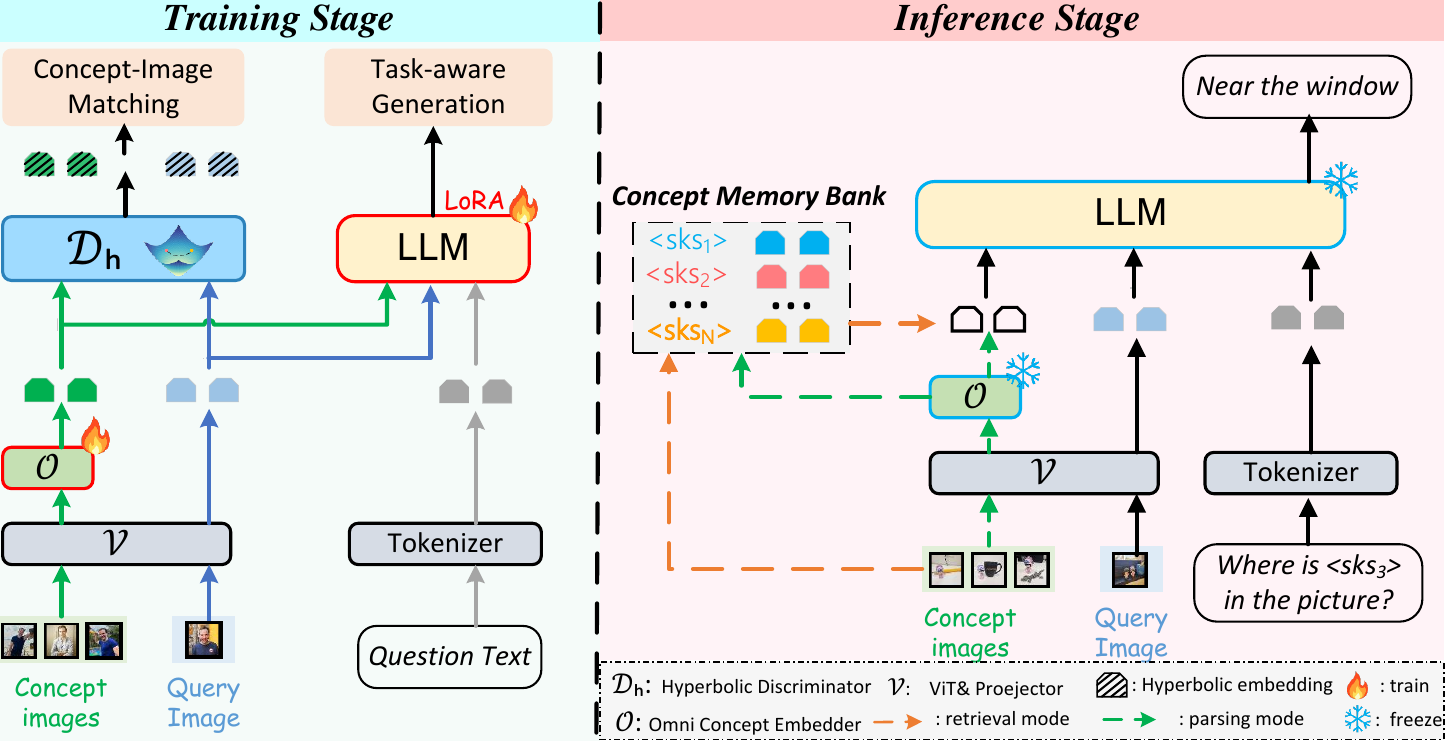}  
    \caption{The whole pipeline of Online-PVLM for the training stage and inference stage.}
    \label{fig:data_flow}
\end{figure*}
For the online concept learning task, we define a concept $c_i$ as a specific object or person. The task input consists of three components: a set of $n$ concept representation images $\mathcal{R}_i = \{ R_i^{(1)}, R_i^{(2)}, \dots, R_i^{(n)} \}$ corresponding to a user-specific concept $c_i$, a query image $Q$ and a set of $m$ natural language questions $\{ q_1, q_2, \dots, q_m \}$ related to $Q$. Each representation image $R_i^{(j)}$ serves as a visual reference that grounds the concept $c_i$. The goal is to enable a VLM $\mathcal{M}$ to process $Q$, $\mathcal{R}_i$, and the question set, and output corresponding answers $A = \{ a_1, a_2, \dots, a_m \}$. 

To address a personalized task for $c_i$, we first generate the corresponding concept embedding $z_i$. Based on this embedding, and in accordance with the methodology described in \citet{nguyen2024yo}, we construct a soft prompt for a new concept in the following manner:

\noindent
\texttt{``<sks\textsubscript{i}> is \textcolor{blue}{<embed.>\textsubscript{1}}}\texttt{\textcolor{red}{<embed.>\textsubscript{2}}}\texttt{\textcolor{teal}{...<embed.>\textsubscript{k}}.''}.

Here, \texttt{<sks\textsubscript{i}>} is an text identifier for concept $c_i$, and the sequence $\{ \langle \text{embed.}_i \rangle \}_{i=1}^k$ is derived from the concept embedding $z_i$. This soft prompt is further concatenated with the normal tokens of the query image $Q$ and the question $q_i$, and the model is trained to generate the corresponding answer.
\subsection{Online-PVLM: A Personalized VLM with Online Concept Learning}

\noindent\textbf{Overview.} Our proposed method, Online-PVLM consists of three key components: the \encoder{} $\mathcal{O}$ for on-the-fly concept embedding, the hyperbolic discrimination module $\mathcal{D}_h$ for structured representation learning, and the LoRA-based VLM $\mathcal{M}$ for instruction following, as illustrated in Figure~\ref{fig:data_flow}.
The training phase of Online-PVLM can be divided into three stages: (1) Concept Embedding Generation (2) Hyperbolic Discrimination Learning, and (3) Joint Training with generative loss.
\vspace{1em}

\noindent\textbf{Concept Embedding Generation.} 
Unlike prior concept learning methods that require training concept-specific embeddings, we demonstrate that a LoRA-based model already exhibits strong capability in handling meta information (as verified in Section~\ref{subsection:novel_concept_test}). Therefore, we only require a lightweight concept embedder to extract omni-level features from visual encoders, which we refer to as \encoder.

From a set of concept images $\mathcal{R}_i$, the \encoder{} $\mathcal{O}$ extracts its instance-normalized ViT features, applies mean pooling, and projects them through an MLP to generate a compact concept embedding $z_i$. This embedding serves as a personalized visual representation of concept $c_i$.

Given $n$ concept images for concept $c_i$, denoted as $R_i \in \mathbb{Z}^{n \times 3 \times H \times W}$, 
the concept embedding for $c_i$ is formulated as:
\begin{equation}
z_i = 
\mathrm{MLP}\!\Bigl(
\frac{1}{n}\sum_{j=1}^{n} 
\mathrm{IN}\bigl(\mathcal{V}(R_i^{(j)})\bigr)
\Bigr),
\end{equation}
where $\mathcal{V}(\cdot)$ denotes the ViT feature extractor with multimodal projector, and $\mathrm{IN}(\cdot)$ applies instance normalization to each feature. 
\vspace{1em}

\noindent\textbf{Hyperbolic Discrimination Learning.}
Projecting representations into hyperbolic geometry is effective in capturing semantic structures in computer vision tasks \citep{li-etal-2022-hypoformer, mettes2023hyperbolicdeeplearningcomputer}. Inspired by this, we project both the query image and concept embeddings into a shared Poincaré ball model \citep{DBLP:journals/corr/NickelK17} and introduce a Hyperbolic Discriminator $\mathcal{D}_h$ to determine whether they refer to the same concept. To supervise this alignment, we adopt a margin-based discriminative loss $\mathcal{L}_{\text{disc}}$ based on hyperbolic distance, which can be formulated as:
\begin{equation}
\begin{aligned}
& \boldsymbol{t}=\mathcal{V}(Q), \ \ \ \ \mathcal{L}_{\text{disc}}
=\mathbb{E}_{\bigl(\mathcal{D}_h(\boldsymbol{t}),\,\mathcal{D}_h(\boldsymbol{z}_i),\,y\bigr)}
\!\bigl[\ell(y)\bigr], \\
& \ell(y)=
\begin{cases}
d_{\text{hypo}}^{\,2}, & y=1,\\[4pt]
\bigl[\max\!\bigl(0,\,\text{margin}-d_{\text{hypo}}\bigr)\bigr]^{2}, & y=0,
\end{cases}
\end{aligned}
\end{equation}
where $Q$ is the input query image, $d_{\text{hypo}}$ denotes the hyperbolic distance, and $y\in\{0,1\}$ is the pair label where 1 representing the same concept.
\vspace{1em}

\noindent\textbf{Joint Training with Generative Loss.} 
To ensure the model fully interprets the injected concept embeddings, we augment its language layers with lightweight LoRA adapters. During training, we jointly minimize the model’s intrinsic auto-regressive loss $\mathcal{L}_{\text{ans}}$ and the discrimination loss $\mathcal{L}_{\text{disc}}$ introduced earlier, allowing both objectives to guide parameter updates.
\begin{equation}
\begin{aligned}
& \mathcal{L}(\theta_{e},\theta_{m})
      \;=\;
      \mathcal{L}_{\text{ans}}(\theta_{e},\theta_{m})
      \;+\;
      \lambda\,\mathcal{L}_{\text{disc}}(\theta_{e}), \\
& \mathcal{L}_{\text{ans}}(\theta_{e},\theta_{m})
=\mathbb{E}
\bigl[-\log P_{\theta_{m}}\!\bigl(A \mid Q,\mathcal{R}_i,\{q_i\};\theta_{e}\bigr)\bigr],
\end{aligned}
\end{equation}
where
$\theta_{e}$ represents parameters of the \encoder,
$\theta_{m}$ represents parameters of the VLM, including the LoRA adapters, and $\lambda$ is a regularization term that balances the two losses.

\subsection{Inference Phase}
\label{subsection:inference}
\begin{figure*}[ht!]  
    \centering
    \includegraphics[width=\textwidth]{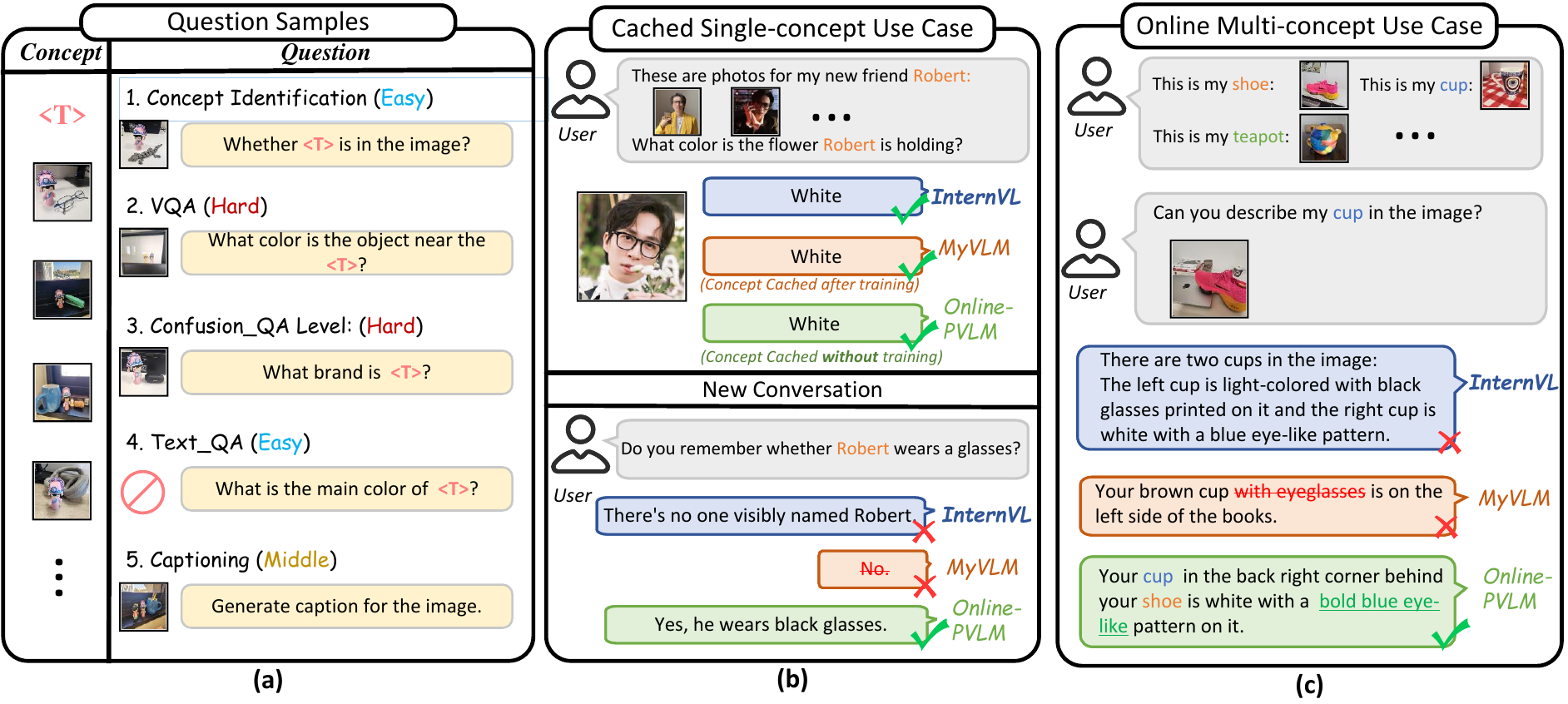}
    \caption{Illustration of question types and use cases for personalized VLMs. \textbf{(a)} Five types of concept-related tasks with varying difficulty levels. \textbf{(b)} Cached single-concept use case showcasing model performance when the user asks follow-up questions based on previously introduced concepts. \textbf{(c)} Online multi-concept use case demonstrating the model’s ability to learn multiple user-provided concept entities on the fly and answer related queries. Both \textbf{(b)} and \textbf{(c)} highlight the effectiveness of Online-PVLM across different application settings.}
    \label{fig:benchmark_demo}
\end{figure*}

During the inference phase, we provide two complementary, training‑free inference modes—retrieval and parsing—to handle previously seen and novel concepts, respectively. The sample use cases for these two modes are illustrated in Figure~\ref{fig:benchmark_demo}(b) and (c).

\textbf{Parsing Mode.}
When a query involves an unseen concept $S_{\text{new}}$, the user provides a small set of reference images $\{R^{(j)}\}_{j=1}^{k}$, which are passed once through the frozen \encoder{} to obtain a new embedding $z_{\text{new}}$. This embedding is immediately used for inference and then cached in the \textit{concept memory bank} for future retrieval. Unlike traditional test-time training methods \citep{alaluf2024myvlm, nguyen2024yo}, this parsing mode supports on-the-fly adaptation to open-vocabulary concepts without gradient updates, while enabling efficient retrieval for subsequent queries. 
\vspace{1em}

\textbf{Retrieval Mode.}
If the personalized embedding $z_i$ for a concept is already cached in the \textit{concept memory bank}, it can be retrieved using the identifier $S_i$ in $O(1)$ time during inference. Unlike traditional online linear methods \citep{pi2024personalized}, which do not incorporate explicit concept embeddings, this retrieval-based approach removes the need for additional reference images and encoder computation, significantly reducing inference latency and energy consumption in practical applications.

%% file: sec/4_dataset.tex
\section{OP-Eval Dataset}
\label{sec:dataset}

Although existing datasets target personalization in VLMs \citep{alaluf2024myvlm, nguyen2024yo, an2024mc}, they mainly focus on constrained settings with ample supervision or limited concept diversity, overlooking key traits of real-world online use—such as handling unseen concepts, cross-concept queries, and sparse data. To bridge this gap, we dedicate substantial effort to cleaning, expanding, augmenting, and manually annotating existing datasets from \citet{pi2024personalized}, \citet{alaluf2024myvlm}, and \citet{nguyen2024yo}, resulting in OP-Eval—a comprehensive benchmark for online concept learning with large-scale and diverse concept–task configurations. We provide a brief overview of the dataset construction pipeline below.

\input{tables/dataset_comparison}

    \noindent \textbf{1. Image Filtering and Partitioning.} Data authenticity and consistency are critical for real-world online applications. To this end, we curate data through a two-stage pipeline. First, we remove low-quality, synthetic, and ambiguous images to ensure dataset reliability. Then, we rank the remaining images by entity prominence, assigning the top-ranked samples to the concept set and the rest to the test set. This process produces a realistic and challenging benchmark.
    
    \noindent \textbf{2. Task Formulation.} Handling diverse downstream tasks is a basic requirement for online concept learning. We design three per-image tasks: Concept Identification, Concept QA, and Concept-based Captioning. Furthermore, Concept QA includes Text QA, Visual QA, and Confusion QA, targeting feature extraction, grounding, and hallucination resistance.
This design enables a comprehensive evaluation of concept understanding.
    
    \noindent \textbf{3. Expanding to Multi-Concept Protocol.} Previous works \citep{nguyen2024yo, alaluf2024myvlm} have focused on single-concept understanding, but real-world scenarios often involve multiple concepts queries. To address this, we extend the standard single-concept protocol by introducing a multi-concept evaluation, where models must recognize and reason about multiple concepts within the same image. Both Protocols share identical question types, enabling direct performance comparison.
    
    \noindent \textbf{4. Data Augmentation and Classification.} We construct question cases by sampling concept images and test images from their respective pools, ensuring scalable and realistic coverage. Questions are further grouped into three difficulty levels based on image prominence, visual similarity, and question complexity.

The resulting OP-Eval comprises 1,292 unique concepts with approximately 3,000 images and 30,000 question cases. It features three core task types: concept identification, question answering (QA), and caption generation. Figure~\ref{fig:benchmark_demo}(a) illustrates examples of different concept types with difficulty level.

%% file: tables/dataset_comparison.tex
\begin{table}[t]        
  \centering             
  \scalebox{0.65}{       
    \begin{tabular}{lccc}
      \toprule
      \textbf{Dataset}  & \textbf{Q. Type} & \textbf{C. Type} & \textbf{C. Num}\\
      \midrule
      MyVLM \citep{alaluf2024myvlm} & G        & Object            & 29\\
      Yo'LLAVA \citep{nguyen2024yo} & Q, I     & People, Object    & 40\\
      PVIT \citep{pi2024personalized}& Q, I, G & People            & 79\,510\\
      MC-LLAVA \citep{an2024mc}      & Q, I, G & People, Object    & 118\\
      OP‑Eval (Ours)                & Q, I, G & People, Object    & 1\,292\\
      \bottomrule
    \end{tabular}
  }
  \caption{\small \textbf{Dataset Comparison.} "\textbf{Q. Type}" means the question type; "\textbf{C. Type}" represents the concept type, and "\textbf{C. Num}" means the number of concepts in the dataset. "Q", "I", and "G" denote QA, Identification, and Captioning.}
  \label{tab:datasets_comparison}
  \vspace{-10pt}
\end{table}

%% file: sec/5_experiment.tex
\section{Experiments}
\input{tables/novel_concept_test}

\input{tables/cached_concept_test}

As described in Section~\ref{subsection:inference}, Online-PVLM can handle two scenarios at test time for unseen concepts and cached concept embeddings.
To comprehensively evaluate the performance of Online-PVLM in these two scenarios, we design two complementary evaluation settings: the Novel Concept Test, which assesses the model's capability to handle previously unseen concepts, and the Cached Concept Test, which measures performance on concepts whose embedding have been encountered and cached by the \textit{concept memory bank}. 

\subsection{Experiment Setup}
\noindent\textbf{Dataset.}
We evaluate both settings using OP-Eval (Section~\ref{sec:dataset}), which comprises concept-level image sets and corresponding questions spanning three task types—concept identification, concept question answering, and concept-aware captioning—under two protocols: single-concept and multi-concept. As no existing dataset supports both evaluation scenarios (see Table~\ref{tab:datasets_comparison}), we additionally employ P-Bench \citep{pi2024personalized} for the Novel Concept Test and MyVLM \citep{alaluf2024myvlm} for the Cached Concept Test. 

P-Bench is an online learning benchmark containing 79,510 person-specific concepts, each associated with multiple-choice VQA and captioning questions. We only evaluate the VQA subset here\footnote{Only the VQA test set in P-Bench is publicly released.}. The MyVLM dataset contains 29 object-level concepts, each paired with identification and captioning tasks.


\vspace{1em}

\noindent\textbf{Training.} Unless stated otherwise, we use AdamW \citep{kingma2017adammethodstochasticoptimization} with a 1e-5 learning rate, and the number of concept embedding tokens is 256. After extensive experiments, we use InternVL2.5-8B \citep{chen2024internvl} as our base model for its strong baseline capabilities. The entire training was conducted on 8 A100 GPUs with 80GB of memory for 6 epochs, which lasted for 4 hours. 

\vspace{1em}

\noindent\textbf{Baselines.}
For the Novel Concept Test, we implement two baselines following \citet{nguyen2024yo}: \textbf{(1) Vanilla VLM}, a standard vision-language model without personalization, as mentioned previously, we adopt InternVL-7B. \textbf{(2) Prompt-Augmented VLM} simulates personalization by injecting a textual description of the target concept (e.g., "a red cup with a floral pattern") into the prompt. Each description, consisting of approximately 30 words, is generated by GPT-4o and verified by human annotators. Additionally, we consider two online training methods: \textbf{(3) PVIT}~\citep{pi2024personalized}: a visual instruction tuning approach without concept embeddings.
\textbf{(4) Online-PVLM w/o LoRA}: a variant of our model that removes the LoRA module.

For the Cached Concept Test, we include two concept embedding learning methods. \textbf{(1) MyVLM}\citep{alaluf2024myvlm}: an offline concept learning method that learns concept representations and concatenates them with test image features.
\textbf{(2) Yo’LLAVA}\citep{nguyen2024yo}: an offline concept learning approach that directly trains embedding tokens for each concept.

\vspace{1em}

\noindent\textbf{Metrics.}
We evaluate OP-Eval following the protocols outlined in \citet{nguyen2024yo} and \citet{alaluf2024myvlm}. For concept identification and question answering, we report accuracy, while the captioning task is assessed using recall and text similarity metrics. In the case of P-Bench, we adopt accuracy as the primary metric for evaluating VQA performance. As for the MyVLM dataset, we follow the original definitions in \citet{nguyen2024yo} and report both accuracy and recall.

 \subsection{Novel Concept Test}
 \label{subsection:novel_concept_test}
The Novel Concept Test evaluates the model’s robustness and generalization to unseen concepts. For OP-Eval, we train on 683 concepts and test on the remaining 608 concepts, and we test 341 concepts in the P-Bench test set.

Unlike baseline methods that directly use raw images to refer to a concept, Online-PVLM encodes concept images into personalized embeddings via its \encoder. Table~\ref{tab:novel_concept_test} presents the evaluation results.

In both concept identification and concept QA tasks emphasizing novel concept recognition and visual grounding, Online-PVLM consistently outperforms baseline models. As anticipated, the vanilla baseline performs near chance level on concept identification due to its inability to recognize novel entities, though it achieves moderately better QA scores, likely owing to the presence of prominent visual cues. The prompt-augmented baseline offers slight improvements but remains inferior to training-based approaches. Online-PVLM considerably outperforms the PVIT method for trainable approaches, even when utilizing fewer input tokens—specifically, concept embeddings rather than multiple images. This leads to a reduction in inference time while simultaneously improving performance.

For the concept-aware captioning task, we evaluate model performance on the downstream objective. The training-free method yields a low recall score, indicating its failure to generate concept identifiers. Online-PVLM occasionally misidentifies concepts in its generated captions. This may be attributed to the discriminative objective used during training, which improves concept grounding but may adversely affect generative capabilities.

Notably, when comparing PVIT, which includes only the LoRA module, with a variant of online-PVLM, which omits LoRA and utilizes only the \encoder{}, it is evident that the former outperforms the latter. This underscores the effectiveness of \textit{LoRA as an online learner}, as previously noted. By combining both approaches, online-PVLM not only enables on-the-fly concept embedding generation for each method but also achieves superior overall performance.

 \subsection{Cached Concept Test}

For the Cached Concept Test evaluation, we assess 29 object-level concepts from the MyVLM dataset alongside a randomly sampled subset of 1,000 concepts from OP-Eval. The adaptation of OP-Eval to this setting involves three key steps:
(1) \textbf{Concept Images Retrieval.} We follow \citet{alaluf2024myvlm} and extract up to four images per concept from the concept image pool for embedding construction.
(2) \textbf{Test Images Partition.} To prevent information leakage, we split test images and their corresponding questions into disjoint training and test sets, ensuring that concept-specific test questions remain unseen during training. Regarding concept embedding generation, for MyVLM and Yo’LLAVA, the learned embeddings are cached and directly used during inference. For Online-PVLM, we employ the trained \encoder{} to generate embeddings after training.

\begin{figure*}[ht!]  
    \centering
    \includegraphics[width=\linewidth]{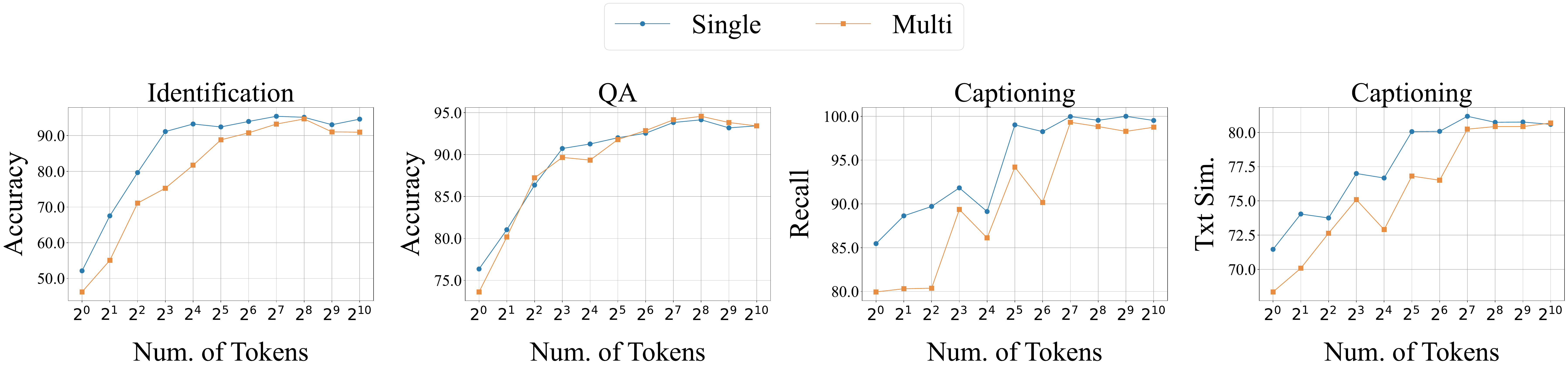}  
    \caption{Ablation study on the token number of personalized concept embedding.}
    \label{fig:ablation_concept_token_num}
    \vspace{-10pt}
\end{figure*}

Table~\ref{tab:cached_concept_test} summarizes the results. On the MyVLM dataset, Online-PVLM demonstrates strong performance in the QA task. Its recall score is 0.69 points below perfect accuracy, further supporting the conclusion drawn from the Novel Concept Test.

For OP-Eval, MyVLM and Yo’LLAVA fail to make progress due to two key limitations: (1) their requirements for concept-specific training become computationally prohibitive at scale (1,000 concepts), and (2) their dependence on extensive negative samples is incompatible with OP-Eval's more realistic, resource-constrained deployment scenario. Online-PVLM demonstrates performance gains over the Novel Concept Test, potentially attributable to the in-domain evaluation.

%% file: tables/novel_concept_test.tex
\begin{table*}[ht]
\centering
\begin{minipage}[b]{0.66\linewidth}
\scalebox{0.63}{  
\begin{tabular}{l|cc|ll|ll|ll|c}
\toprule
\multicolumn{1}{c|}{\textbf{Dataset}} & \multicolumn{8}{c|}{{\textbf{OP-Eval}}} & \multicolumn{1}{c}{{\textbf{P-Bench}}} \\

\cmidrule(lr){1-1}
\cmidrule(lr){2-9}
\cmidrule(lr){10-10}

\multicolumn{1}{c|}{\textbf{Task Type}} & \multicolumn{2}{c}{{\textbf{Identification}}} & \multicolumn{2}{c}{{\textbf{QA}}} & \multicolumn{4}{c|}{\textbf{Captioning}} & \multicolumn{1}{c}{\textbf{QA}}\\

\cmidrule(lr){1-1}
\cmidrule(lr){2-3}
\cmidrule(lr){4-5}
\cmidrule(lr){6-9}
\cmidrule(lr){10-10}

\multicolumn{1}{c}{\multirow{2}{*}{\textbf{Q. Setting}}} & \multicolumn{2}{c}{{\textbf{Accuracy}}} & \multicolumn{2}{c}{{\textbf{Accuracy}}} & \multicolumn{2}{c}{\textbf{Recall}} & \multicolumn{2}{c|}{ \textbf{Txt Sim.}} & \multicolumn{1}{c}{ \textbf{Accuracy}}\\
& \multicolumn{1}{c}{Single} & \multicolumn{1}{c}{Multi} & \multicolumn{1}{c}{Single} & \multicolumn{1}{c}{Multi} & \multicolumn{1}{c}{Single} & \multicolumn{1}{c}{Multi} & \multicolumn{1}{c}{Single} & \multicolumn{1}{c|}{Multi} & \multicolumn{1}{c}{Single}\\

\midrule
InternVL \citep{chen2024internvl} & 57.86 & 53.56 & 89.36 & 83.44 & 17.38 & 20.91 & 56.01 & 55.83 & 56.43\\
InternVL* & 73.52 & 71.64 & 91.23 & 86.72 & 18.62 & 22.34 & 64.32 & 65.48 & 61.84\\

Online-PVLM (w/o LoRA) & 86.43 & 85.15 & 91.31 & 91.07 & 98.68 & 97.46 & 79.28 & 79.36 & 94.54\\

PVIT\citep{pi2024personalized} & \underline{90.59} & \underline{90.74} & \underline{92.65} & \underline{92.76} & \textbf{100.0} & \textbf{99.75} & \textbf{82.77} & \textbf{82.55} & \underline{96.32}\\

\textbf{Online-PVLM (Ours)} & \textbf{95.13} & \textbf{94.66} & \textbf{94.14} & \textbf{94.56} & \underline{99.54} & \underline{98.83} & \underline{80.74} & \underline{80.43} & \textbf{97.41}\\
\bottomrule
\end{tabular}
}
\caption{Performance of models under \textbf{Novel Concept Test}. The \textbf{best} and \underline{second-best} performances in each column are highlighted. Txt Sim. represents text similarity. * refers to prompt augmented method.}
\label{tab:novel_concept_test}
\end{minipage}
\hfill
\begin{minipage}[t]{0.3\linewidth}
\vspace{-50mm}
  \raggedleft
  \includegraphics[width=1.0\linewidth]{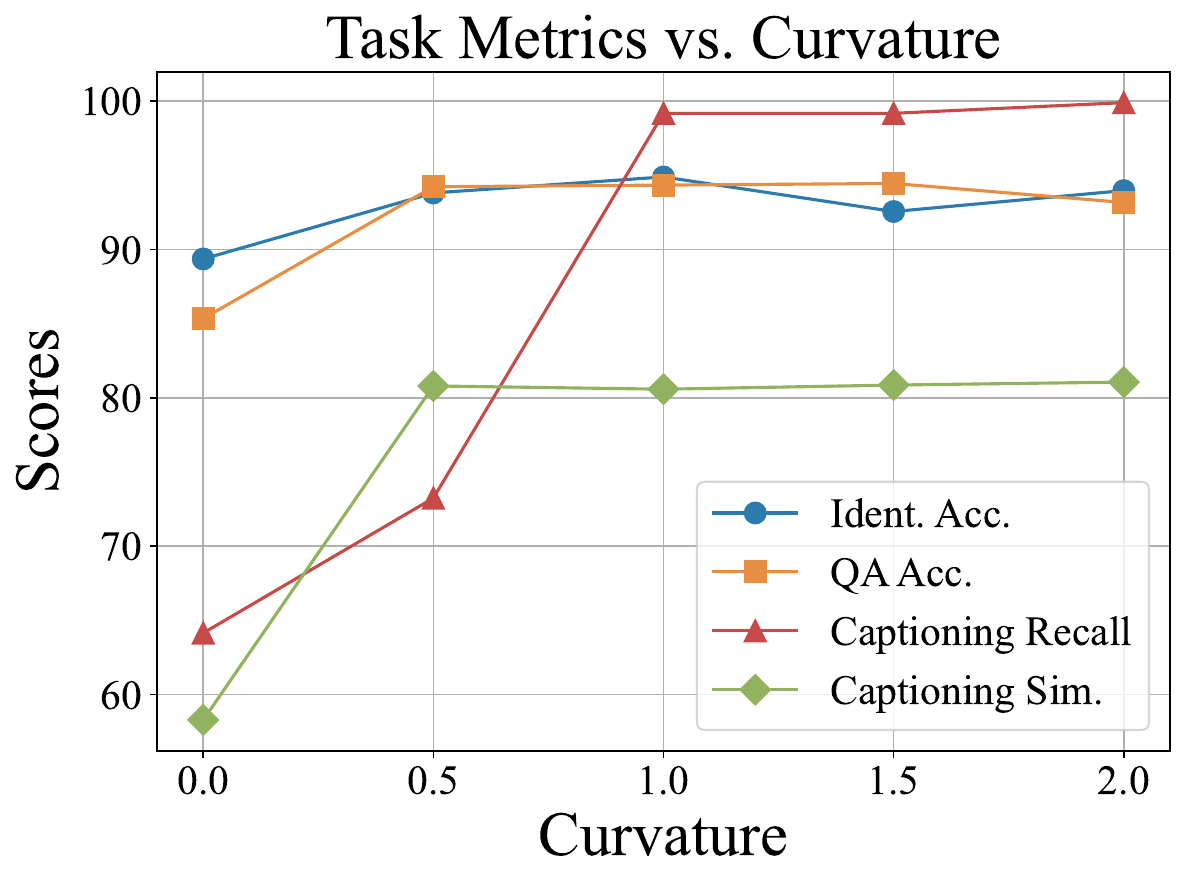}  
  \vspace{-6.5mm}
  \captionof{figure}{Ablation study on the hyperbolic curvature value.} 
  
  \label{fig:ablation_hypo_value}
\end{minipage}
\end{table*}

%% file: tables/cached_concept_test.tex
\begin{table}[h]
\begin{minipage}[b]{\linewidth}  
\scalebox{0.8}{  
    \begin{tabular}{l|cc}
    \toprule
    {\textbf{Model Type}} & \textbf{Acc.} & \textbf{Recall} \\
    \cmidrule(lr){1-3}      
    \multicolumn{3}{c}{\cellcolor{gray!20}  Concept Num. = 29 \small{\textit{(on MyVLM dataset)}}} \\
    MyVLM \citep{alaluf2024myvlm}& 93.8 & 96.0 \\
    Yo'LLAVA  \citep{nguyen2024yo}  & \underline{96.40} & \textbf{100.0} \\
    \textbf{Online-PVLM (Ours)}  & \textbf{96.75} & \underline{99.31} \\
    \midrule
    \multicolumn{3}{c}{\cellcolor{gray!20}  Concept Num. = 1000 \small{\textit{(on OP-Eval)}} } \\
    MyVLM   & Infeasible† & Infeasible† \\
    Yo'LLAVA  & Infeasible† & Infeasible† \\
    \textbf{Online-PVLM (Ours)}  & \textbf{95.34} & \textbf{99.64} \\
    \bottomrule
    \end{tabular}
}
\end{minipage}
\caption{Performance of models under the \textbf{Cached Concept Test}. The accuracy scores (Acc.) for the identification task and recall scores for the captioning task are presented. "Infeasible†" indicates that the method does not support this setting due to architectural constraints.}
\label{tab:cached_concept_test}
\vspace{-10pt}
\end{table}

%% file: sec/6_ablation.tex
\section{Ablation study}


\paragraph{Module Impact on Performance.} As shown in Table~\ref{tab:ablation_module}, we conduct an incremental ablation to assess the contribution of three modules. Adding instance normalization leads to a noticeable improvement in both identification and VQA tasks, which is due to its stabilizing effect on concept visual representations. Introducing discriminative loss further refines the supervision, helping the model better capture concept-specific distinctions. The most substantial gain comes from incorporating our hyperbolic representation (HyperRep), which enhances the model’s ability to encode concept semantics in a geometrically meaningful space. This progressive enhancement highlights the complementary roles of normalization, discriminative guidance, and geometric modeling in building Online-PVLM.
\input{tables/ablation_module}

\paragraph{Number of Embedding Vectors.}
We vary the number of embedding vectors $k$ of the concept embedding from 1 to 1024 in exponential steps. As shown in Figure \ref{fig:ablation_concept_token_num}, performance is poor when $k < 16$, improves steadily as $k$ increases and begins to plateau around $k = 256$. This trend is consistent across all tasks. In the Identification and QA tasks, accuracy increases rapidly with more tokens and stabilizes beyond $k = 256$. In the Caption task, both recall and text similarity improve overall, but the multi-concept setting shows greater variance and lags behind the single-concept variant, especially at smaller vector lengths—likely due to the increased difficulty of encoding multiple concepts. Based on this analysis, we choose $k = 256$ as the default setting to balance performance and efficiency.

\paragraph{Curvature of the Poincaré ball.}
We explore the effect of varying hyperbolic curvature values in the discriminator. As shown in Figure~\ref{fig:ablation_hypo_value}, all task metrics—including identification accuracy, QA accuracy, and both recall and text similarity for generation—improve substantially when the curvature increases from 0 (Euclidean space) to 1.0. This highlights the effectiveness of hyperbolic geometry in capturing concept semantics relationships. Beyond 1.0, performance quickly saturates and remains stable, indicating that extreme curvature offers no further benefit. Notably, QA accuracy reaches its peak at curvature = 1.0, while other metrics remain relatively unchanged when the curvature is set to 1.0 or higher. Based on these observations, we adopt a curvature value of 1.0 as the default setting for balancing stability and representational capacity.

%% file: tables/ablation_module.tex
\begin{table}[h]
\centering
\resizebox{0.48\textwidth}{!}{%
\begin{tabular}{l|cc}
\hline
\textbf{Model Variant} & \textbf{Ident. Acc. (\%)} & \textbf{VQA Acc. (\%)} \\
\hline
w/o Instance Norm            & 89.67 & 93.78 \\
+ Instance Norm              & 91.24 & 94.04 \\
+ Discriminative Loss             & 92.74 & 94.21 \\
+ \textbf{HyperRep (Ours)}   & \textbf{94.91} & \textbf{94.35} \\
\hline
\end{tabular}
}
\caption{Ablation Study on modules of Online-PVLM.}
\label{tab:ablation_module}
\vspace{-10pt}
\end{table}

%% file: sec/7_conclusion.tex
\section{Conclusion}
In this paper, we present Online-PVLM, a novel framework for online concept learning in personalized VLMs, which enables dynamic adaptation without test-time training. To support evaluation, we introduce OP-Eval, a large-scale benchmark tailored for this setting. Extensive experiments across multiple tasks validate the effectiveness and scalability of our approach, highlighting its potential for real-world personalized multimodal applications.

%% file: sec/8_limitation.tex
\section{Limitations}
While our method enables effective personalization through concept-specific embeddings, it remains primarily tailored to representing individual entities. This narrow scope limits its applicability to broader forms of personalization, such as modeling abstract concepts, social groups, or activity categories. Defining and representing such high-level concepts in a unified embedding space remains an open challenge. Moreover, current approaches lack mechanisms for modeling the interactions between multiple personalized concepts, which may be essential for more complex reasoning. This highlights a fundamental limitation in extending personalized VLMs beyond entity-level embeddings toward richer and more structured personalization paradigms.

\section{Ethics Statement}
The response generated by the VLM may contain biased sentences, which may offend the readers. This can be attributed to the potential bias of PLMs \citep{gallegos2024biasfairnesslargelanguage, guo2024biaslargelanguagemodels}. These biased sentences do not reflect the views of the authors.

%% file: sec/9_appendix.tex
\section*{Appendix}

\vspace{0.5cm}
\section{Details of Dataset Construction}
\label{appendix: data_construction}
\subsection{Image Filtering Process}
\begin{figure}[h]  
    \centering
    \includegraphics[width=\linewidth]{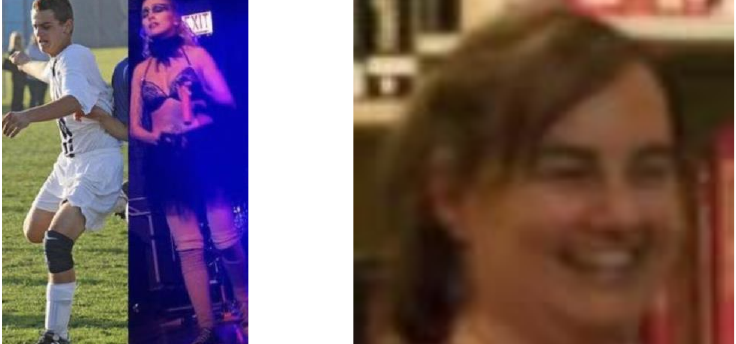}  
    \caption{Examples of low-quality images. The left image is an unnatural composite crop unlikely to occur in real-world scenarios. The right image is excessively blurred, making it difficult to distinguish details.}
    \label{fig:bad_image_case}
\end{figure}

In our dataset, particularly in PVIT, we identified a subset of artificially composed images where multiple concepts are cropped together. To address this, we apply predefined categories, such as "Aug-Sc-2 description" and "Aug-Sc-2 choice", to filter out these instances. For filtering low-quality images, we first utilize the Variance of the Laplacian method \citep{4109404} as an initial filter, followed by human verification to ensure the quality. The problematic image samples are presented in Figure \ref{fig:bad_image_case}.
\subsection{Image Partitioning Process}
\begin{figure}[h]  
    \centering
    \includegraphics[width=\linewidth]{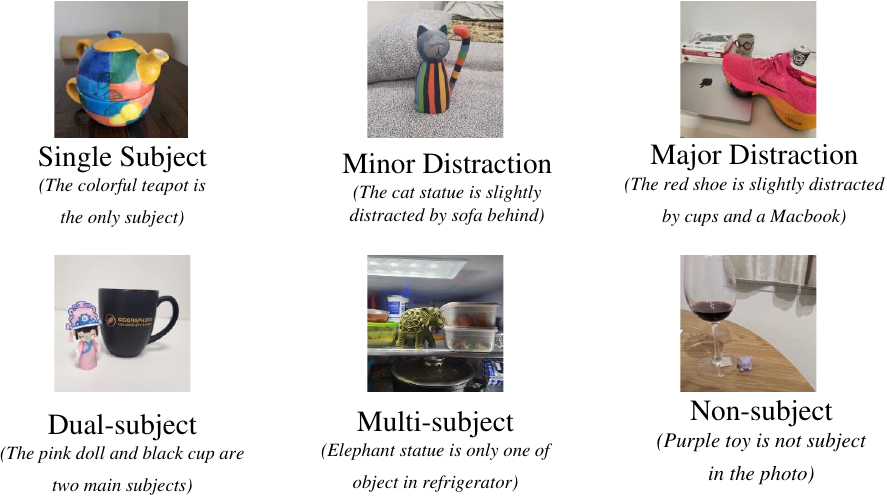}  
    \caption{Examples of six image types categorized by prominence.}
    \label{fig:six_prominent_type}
\end{figure}

To categorize images based on prominence, we employ a heuristic approach that classifies images into six categories: \textit{Single Subject}: target concept is the main subject and no other objects, \textit{Minor Distraction}: target concept is the main subject with minimal visual noise, \textit{Major Distraction}: target concept is the main subject but partially occluded or cluttered, \textit{Dual Subject}: two equally salient subjects are present, \textit{Multi-Subject}: multiple distinct subjects appear, with unclear focus, and \textit{Non-Subject}: target concept is absent or only in the background. This classification process involves evaluations from two human annotators and the GPT-4o-0806 \citep{achiam2023gpt}, with the final category assigned based on the average score. Examples for each category are shown in Figure \ref{fig:six_prominent_type}.
For each concept, we select the top six images as its concept image set (rank from \textbf{Single Subject}, and the left serves as test images. This design ensures that concept images are semantically representative, while test images offer greater variability for robust evaluation.

\subsection{Question Generation Process}
Here are details introductions for each question type:
\begin{itemize} 
    \item \textbf{Concept Identification} involves predicting whether a specified concept is present in a test image. We take the test images selected in the aforementioned step as positive images and introduce negative images scraped from the website by Bing Image Search to form the negative samples.
    \item \textbf{Concept QA} task evaluates the performance of VLMs in providing detailed answers regarding a concept. We categorize this task into 3 types: (1) \textit{Text-based QA} \citep{nguyen2024yo}, where the concept images are provided, query basic details about the concept, such as its color and shape. On the other hand, (2) \textit{VQA} based on test images focuses more on visual grounding information, such as the relative position of the concept and the expressions or states depicted. 
    To assess the robustness of these models, we further propose (3) \textit{Confusion VQA} task. Given a test image, the question aims to deliberately confuse the model by asking questions about features unique to a different object, causing the model to mistakenly identify that object as the concept in question.
    This is essential because we observe that many VLMs have a strong tendency to misinterpret objects as the target concept due to their reliance on superficial features. 
        
    \item \textbf{Concept-based Caption Generation} requires the model to generate a caption that reflects the mentioned concept. The generated caption should focus on describing the image from the perspective of the target concept, highlighting its relevance or prominence in the image, we generate 3 to 5 variants for each test image for diversity and accuracy in concept representation.

\end{itemize}

For normal tasks except for the Confusion QA, we prompt the GPT-4o-0806 model using concept images to generate detailed descriptions of each concept. These descriptions are then combined with the concept images and test images to generate corresponding questions.

For the Confusion QA task, we first use GPT-4o-0806 to generate a description of the test image. This description is then combined with the concept description, the test image description, and representative examples to formulate confusion-based questions, which are generated using the OpenAI o1-12-17 \citep{openai2024o1}.

All generated questions are cross-validated by two human annotators to ensure their quality and consistency.

\subsection{Data Augmentation and Classification}
Once the question tasks and settings are defined, we assemble the full set of question cases. Following the image partitioning process, each concept is associated with a concept image pool and a test image pool. Concept images are randomly drawn from the concept pool for data augmentation, while test images and their corresponding questions—constructed—are systematically sampled from the test pool. This sampling strategy ensures broad coverage and enables a thorough evaluation of the VLM across varied and realistic scenarios. Additionally, to provide a more detailed evaluation, we categorize the questions into three difficulty levels based on image prominence, similarity between test and concept images, and the overall difficulty of the associated questions.

\subsection{Quality Assurance}
The annotation process was carried out by a team of three graduate students, each with over four years of experience in programming and related NLP research. To ensure annotation quality and consistency, all annotations underwent cross-validation among annotators. Furthermore, the results were verified to align closely with outputs generated by large language models (LLMs), ensuring both reliability and coherence with model behavior.
\subsection{Impact of Prominence}
To show the importance of splitting images for prominence, we make an experiment where we use hard-to-type (\textit{Multi-Subject} and \textit{Non-Subject}) as the concept images, and use middle type (\textit{Dual-Subject} and \textit{Major-Distraction}) images for the test. As shown in Table \ref{tab:prominence_check}, using hard-type questions will greatly lower the performance of the models.
\input{tables/appendix/prominence_check}
\label{subsection:prominence}
\subsection{Dataset Statistics}
\label{subsection:dataset_statistics}
\begin{figure}[ht]  
    \centering
    \includegraphics[width=\linewidth]{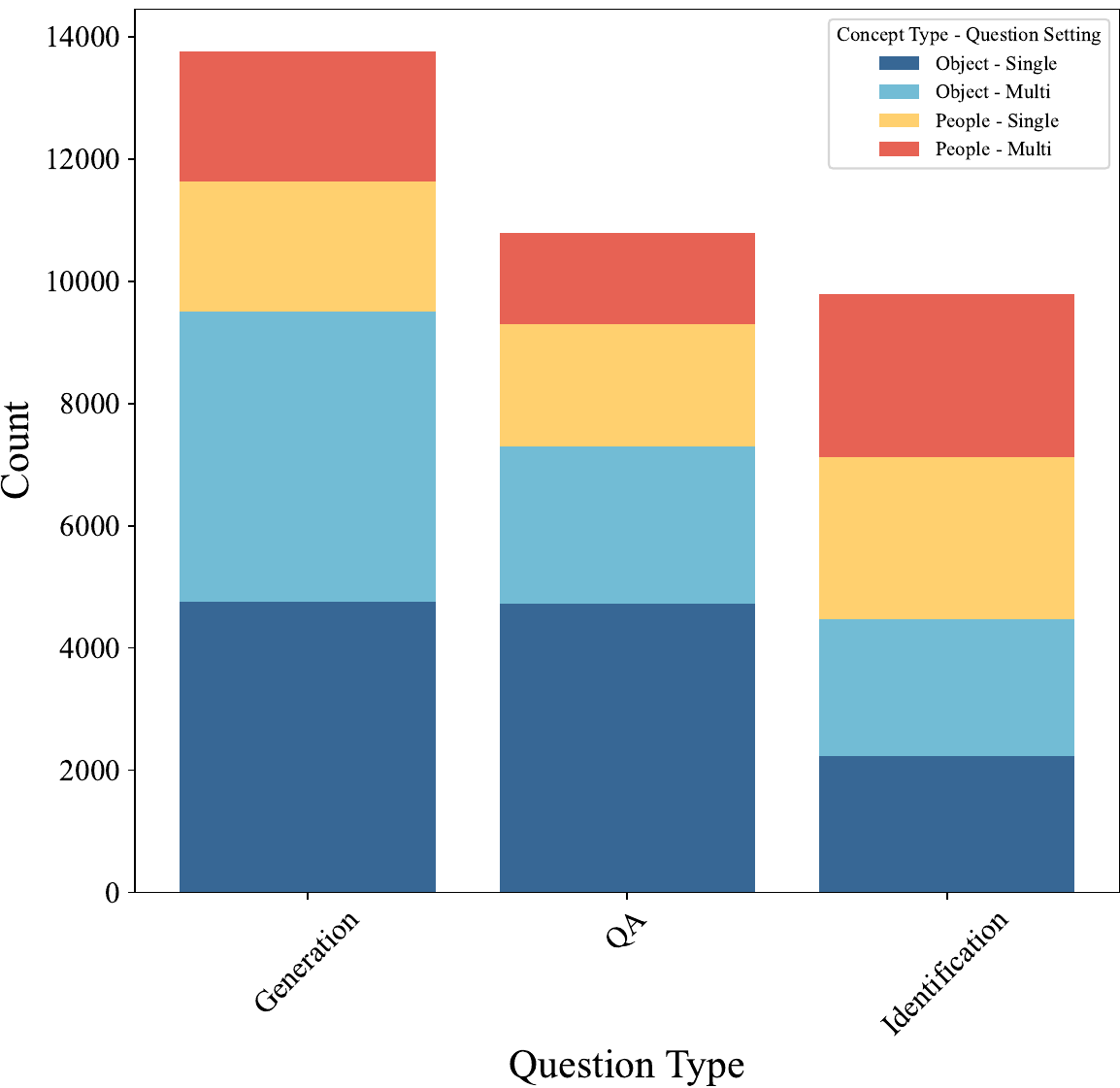}
    \caption{Statistics of OP-Eval.}
    \label{fig:dataset_statistics}
\end{figure}

Figure \ref{fig:dataset_statistics} illustrates the distribution of OP-Eval cases across three question types—Generation, QA, and Identification—and further breaks down the data by concept type (Object vs. People) and question setting (Single vs. Multi concept). 

The Caption Generation category comprises the largest number of samples, totaling nearly 14,000, with a dominant portion derived from object-based concepts. Notably, single-concept object cases form the largest single group within Generation, followed by multi-concept object cases, suggesting the benchmark emphasizes object-centric narrative generation.

The QA (Question Answering) category contains fewer cases than Generation but still surpasses 11,000 instances. Here, both object and people concepts are more evenly represented, and single-concept settings remain predominant. This distribution reflects the importance of structured reasoning over diverse concept types in personalized visual question answering.

The Identification category has the smallest overall sample size, with fewer than 11,000 cases, yet it contains the highest proportion of people-related questions, especially under the multi-concept setting. This indicates a strong focus on evaluating the model's ability to perform fine-grained identification over multiple human-centric concepts, a key aspect of personalized understanding.

Overall, OP-Eval offers a well-balanced benchmark across task types, while intentionally overrepresenting object-centric tasks in generation and human-centric tasks in identification. This distribution design highlights the different capabilities of personalized VLMs and supports comprehensive evaluation across varied scenarios.

\section{Confusion QA}
\begin{figure}[ht!]  
    \centering
    \includegraphics[width=\linewidth]{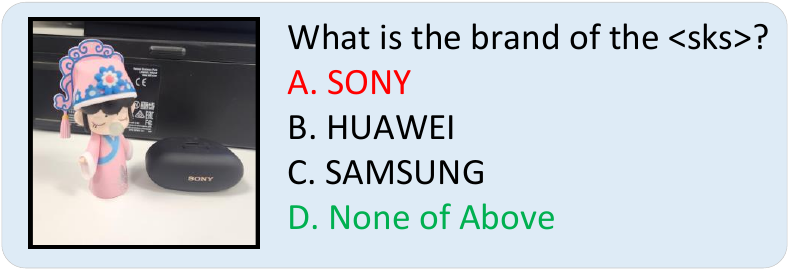}
    \caption{\textbf{An example of Confusion QA.} In this example, the intended concept is the pink doll. Nevertheless, the model incorrectly associates the earphone box with the concept will erroneously chooses “SONY” as the answer.}
    \label{fig:confusion_qa_example}
\end{figure}

To examine the confusion-question benchmark, we sample VQA items from ten concepts and evaluate two conditions: (i) supplying the concept images and (ii) omitting them. As shown in Table \ref{tab:confusion_qa_valid}, a conventional VQA model attains near-full scores on regular VQA even without concept images, whereas its accuracy on confusion questions drops sharply when the images are in the same setting. This contrast indicates that the confusion benchmark effectively mitigates hallucination. Figure \ref{fig:confusion_qa_example} shows an example of the Confusion QA.
\input{tables/appendix/confusion_qa_valid}

\section{Hyperbolic Geometry and Poincaré model}
\label{appendix:hyperbolic}
\noindent\textbf{Hyperbolic Geometry for Deep Learning.}
Hyperbolic space is a non-Euclidean space with constant negative curvature, well-suited for modeling hierarchical or structured data. 
\citet{DBLP:journals/corr/NickelK17} first proposed using the hyperbolic space to learn hierarchical representations of symbolic data such as text and graphs by embedding them into a Poincaré ball. Since then, the use of hyperbolic geometry has been explored in several different applications \citep{li-etal-2022-hypoformer, mettes2023hyperbolicdeeplearningcomputer}.
\vspace{1em}

\noindent\textbf{Hyperbolic Projection for Discriminative Embedding.}
To enhance the discriminative capacity of learned concept embeddings, we project both concept and image features into a hyperbolic space, specifically the Poincaré ball model \citep{DBLP:journals/corr/NickelK17}. The Poincaré ball $\mathbb{D}^n = \{\mathbf{x} \in \mathbb{R}^n : \|\mathbf{x}\| < 1\}$ is a Riemannian manifold with constant negative curvature, which is particularly effective for representing hierarchical and fine-grained data structures.

In this space, the distance between two points $\mathbf{u}, \mathbf{v} \in \mathbb{D}^n$ is defined as:
\begin{equation}
d_{\mathbb{D}}(\mathbf{u}, \mathbf{v}) = \operatorname{arcosh}\left(1 + \frac{2 \|\mathbf{u} - \mathbf{v}\|^2}{(1 - \|\mathbf{u}\|^2)(1 - \|\mathbf{v}\|^2)}\right).
\end{equation}
This formulation leads to a key property: the distance grows exponentially as points approach the boundary of the unit ball, i.e., as $\|\mathbf{x}\| \to 1$. This characteristic makes the Poincaré ball particularly suitable for capturing subtle distinctions between concept embeddings and encourages the separation of semantically different instances. By projecting embeddings into this space, we amplify the discriminative margin between concepts, allowing the model to more effectively distinguish visually similar but semantically distinct entities. This projection is integrated into our contrastive learning framework to improve concept-level representation quality.

\section{Experiment Details}
\label{appendix:implementation_details}
\noindent{\textbf{Model Selection.}}
To determine a suitable backbone for our baseline, we evaluate several mainstream open-source VLMs with model sizes around 8B. As shown in Table \ref{tab:ablation_zero_shot}, InternVL achieves the best overall performance among the candidates. Given the architectural similarities across these models, we expect that Online-PVLM can be seamlessly adapted to other backbones with minimal modifications. Therefore we choose InternVL as our backbone model.
\input{tables/appendix/zero_shot_vlms}
\vspace{1em}

\noindent{\textbf{Metrics.}}
We use \textit{accuracy} as the evaluation metric for both identification and QA tasks, as they are formulated as multiple-choice questions. For caption generation, we report \textit{recall}, which measures whether the concept identifier (e.g., $<\text{sks}_i>$) appears in the generated text. However, \textit{recall} alone may be insufficient, as semantically correct descriptions (e.g., "the dog is in the image") may not include the explicit identifier and thus be incorrectly penalized. To address this, we additionally compute \textit{text similarity} between the generated and reference captions using the Sentence-Transformers model all-MiniLM-L6-v2 \citep{reimers2019sentencebertsentenceembeddingsusing}.

\section{Ablation Studys}
\subsection{Omni Concept Embedder Head}
To explore the most effective design for generating concept embeddings, we evaluate six lightweight encoder head variants within the \encoder. The configurations are as follows:

\begin{itemize}
\item \textit{Direct\_Cross\_Attn}: The $n$ concept features extracted via $\mathcal{V}(\cdot)$ are concatenated and used as keys and values in a cross-attention layer, with the test image serving as the query.

\item \textit{Cross\_Attn}: We apply mean pooling over the $n$ concept features to obtain a single concept representation, which is used as the key, value, and query in a cross-attention layer conditioned on the test image.

\item \textit{Reverse\_Cross\_Attn}: Similar to \textit{Cross\_attn}, but the mean-pooled concept embedding is treated as the query, and the test image features are used as keys and values.

\item \textit{Self\_Attn}: The mean-pooled concept features are used in a self-attention layer to allow internal interaction and refinement.

\item \textit{MLP}: A simple two-layer MLP, where the first layer maps the input from $d_{\text{dim}}$ to $4 \cdot d_{\text{dim}}$, followed by a second layer projecting it back to $d_{\text{dim}}$.

\item \textit{Multi-MLP}: A deeper variant consisting of two stacked MLPs following the same design as above.

\end{itemize}

\begin{figure*}[ht!]  
    \centering
    \includegraphics[width=\linewidth]{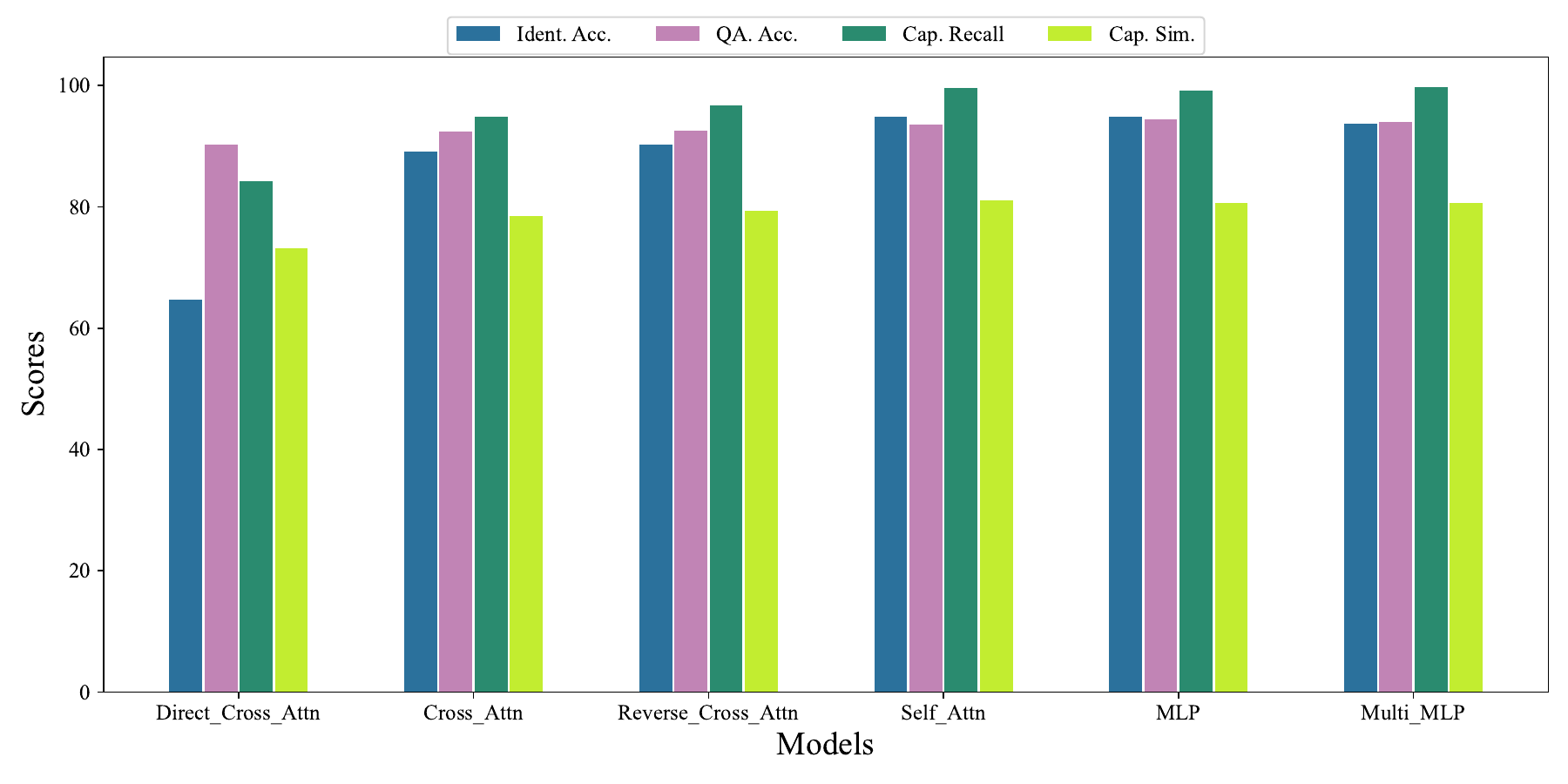}
    \caption{Ablation study on different head types of Omni Concept Embedder.}
    \label{fig:ablation_head_type}
\end{figure*}

The results are summarized in Figure \ref{fig:ablation_head_type}. \textit{Direct\_cross\_attn} performs significantly worse on Identification Accuracy, indicating that directly concatenating raw concept features may introduce noise and hinder effective concept grounding.
\textit{Cross\_attn }and \textit{Reverse\_cross\_attn} achieve balanced performance, with the reverse direction slightly outperforming the forward one, suggesting that using concept features to attend to the test image may better preserve concept semantics.
\textit{Self\_attn} achieves the best results in Identification and QA Accuracy, benefiting from internal refinement among concept features via self-attention.
\textit{MLP} and \textit{Multi\_MLP} also show strong performance, particularly in Caption Recall, indicating that simple feature fusion may suffice for concept representation when pretrained ViT features are strong. Across all metrics, \textit{Self\_attn} and \textit{Multi\_MLP} strike the best balance, showing consistently high scores in both classification and generation tasks. The results suggest that both attention-based and MLP-based heads can effectively model concept embeddings, but self-attention offers better generalization for identification and QA tasks, while MLP variants show advantages in generative tasks.

\begin{figure*}[ht!]  
    \centering
    \includegraphics[width=\linewidth]{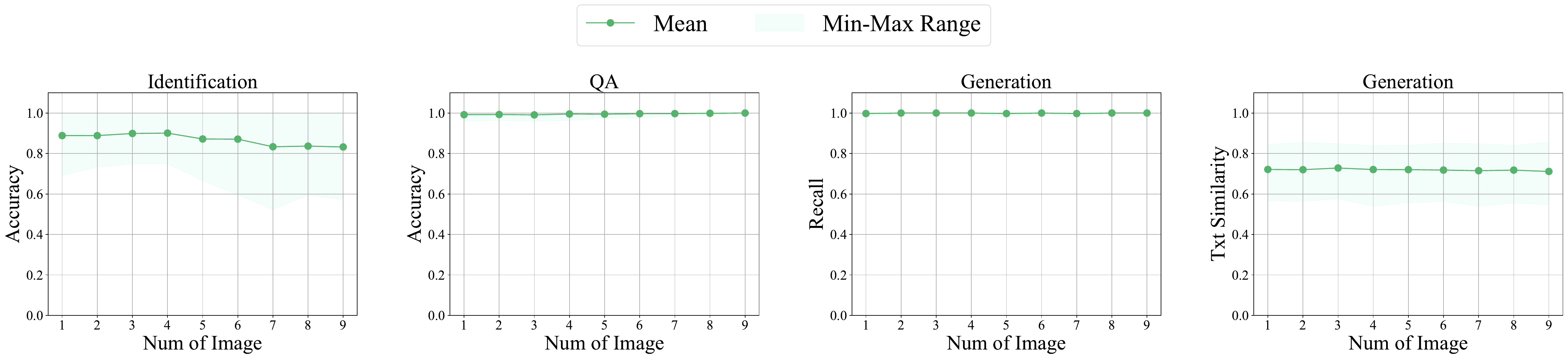}  
    \caption{Ablation study on the number of input images for personalized concept embedding generation.}
    \label{fig:ablation_image_num}
\end{figure*}

\subsection{Number of Input Images.}

We evaluate robustness by varying the number of input concept images during inference from 1 to 9. As shown in Figure \ref{fig:ablation_image_num}, performance remains generally stable. Minor drops are observed in identification and captioning similarity when the number exceeds 6, which demonstrates our strong generalization and adaptation.

\subsection{Computing Cost}
\input{tables/appendix/ablation_computing_cost}
Although Online-PVLM achieves superior performance compared to existing methods, it incurs a higher computational cost during training. As shown in Table~\ref{tab:ablation_computing_costs}, the overall training cost of our approach exceeds that of the LoRA-based baseline and the variant using only the \encoder. This increase is expected, as our full model integrates multiple modules for dynamic personalization, trading off efficiency for improved adaptability and accuracy.

\section{Qualitative Comparison with Baselines}
\label{appendix:qualitative_comparison}
\begin{figure*}[ht!]  
    \centering
    \includegraphics[width=0.9\textwidth]{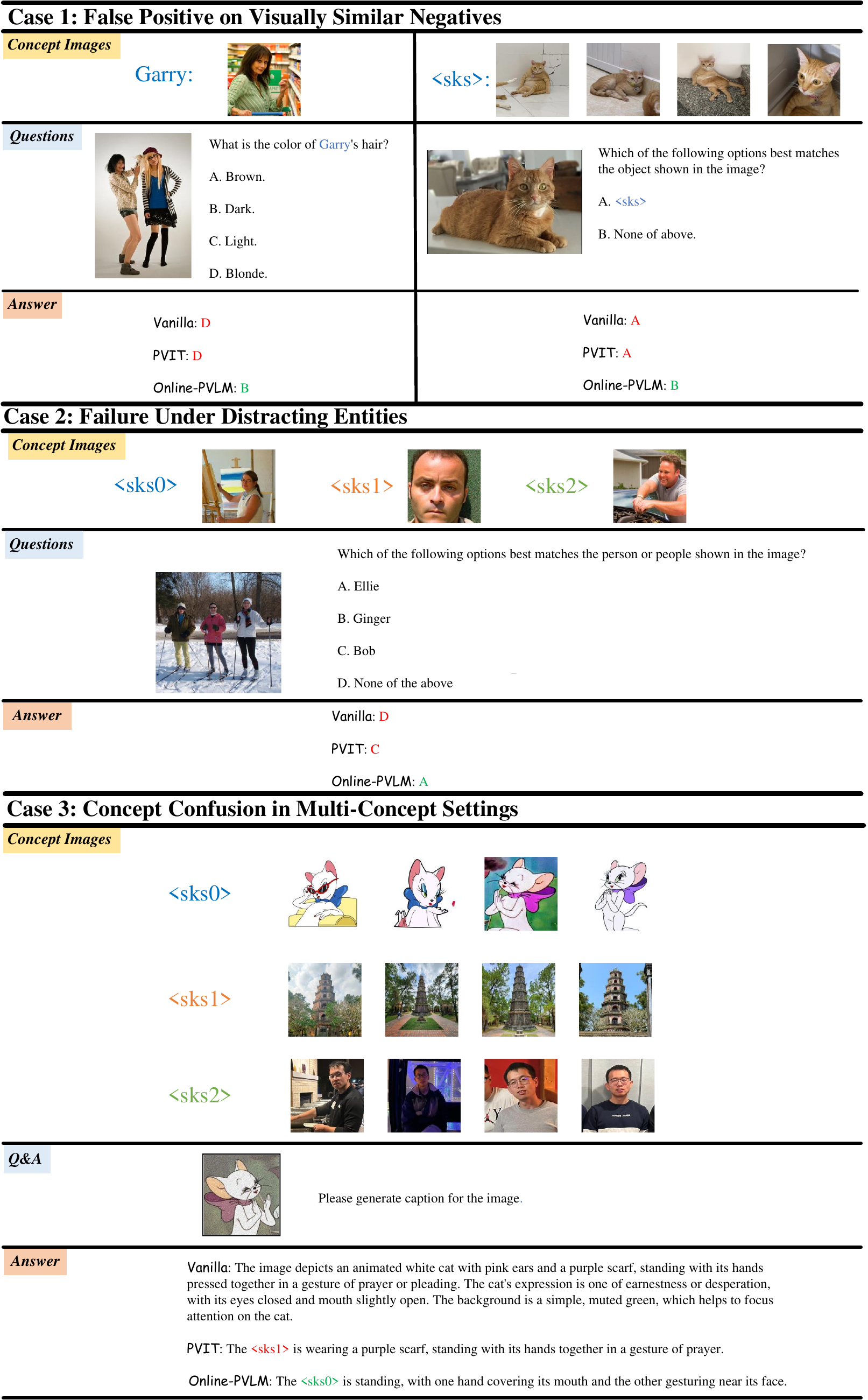}  
    \caption{Qualitative comparison with Vanilla and PVIT methods.}
    \label{fig:qualitative_comparison}
\end{figure*}

Figure \ref{fig:qualitative_comparison} presents representative examples comparing our model with the vanilla baseline and PVIT under both single-concept and multi-concept protocols. We observe three common failure types in the baselines:

\textbf{(1) False Positive on Visually Similar Negatives:} In the single-concept setting (Case 1), both baselines mistakenly identify visually similar but incorrect entities as the target concept, indicating insufficient discrimination.

\textbf{(2) Failure Under Distracting Entities:} When multiple entities appear in the image (Case 2), the models struggle to correctly localize the target concept, suggesting limited robustness to interference.

\textbf{(3) Concept Confusion in Multi-Concept Settings:} In the multi-concept scenario (Case 3), the baselines sometimes recognize relevant concepts but confuse their identities, reflecting instability in fine-grained concept alignment.

These cases highlight the advantages of Online-PVLM in accurately grounding and distinguishing concepts under challenging conditions.

\section{Failure Case Analysis}
\begin{figure*}[ht!]  
    \centering
    \includegraphics[width=0.9\textwidth]{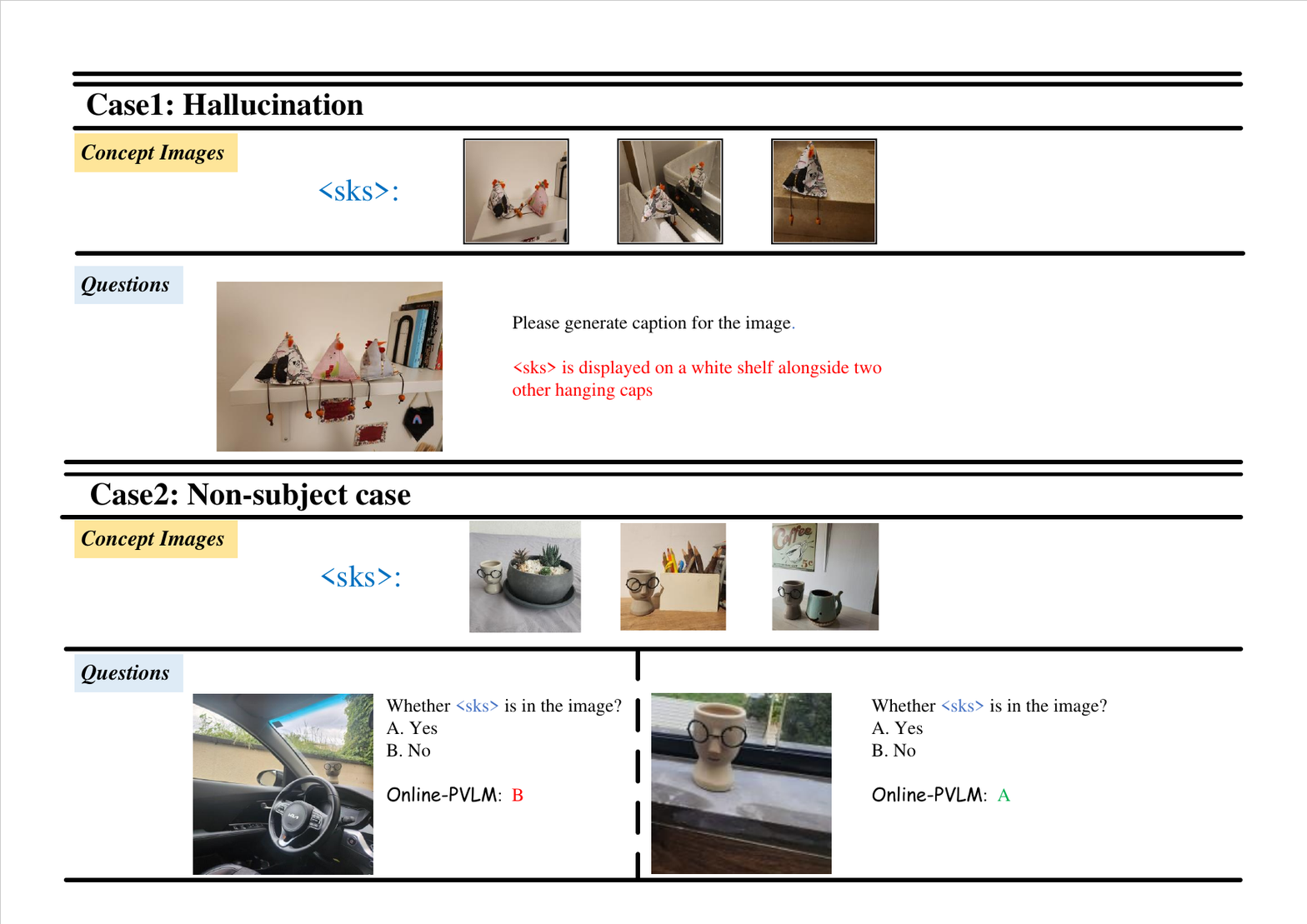}  
    \caption{Failure Case Analysis on Online-PVLM.}
    \label{fig:failure_case_analysis}
\end{figure*}

Although Online-PVLM demonstrates strong overall performance, it still exhibits several limitations. As illustrated in Figure~\ref{fig:failure_case_analysis}, Online-PVLM remains vulnerable to hallucination errors. In Case 1, for example, the model incorrectly identifies the <sks> as a hanging object in the test image, which actually contains six distinct objects. Furthermore, when there is a significant visual discrepancy between the concept images and the test image—such as differences in prominence, lighting, or viewpoint—the model may fail to recognize the primary object. As shown in Case 2, the model can still produce correct predictions under similar lighting and viewpoints, but excessive visual variation or cluttered backgrounds often lead to incorrect answers.

\section{Prompt Format}
We also provide prompt templates used in OP-Eval construction, which are from Figure \ref{tab:general_concept_description} to \ref{tab:ablation_confusion_qa}

\begin{table*}[ht!]
\centering
\begin{minipage}{1.0\textwidth}
\vspace{0mm}    
\centering
\begin{sectionbox}[]{Concept General Description Generation}
    \centering
      \footnotesize
    \begin{tabular}{p{0.97\textwidth} c}
\textbf{\# BACKGROUND}\\
\vspace{1pt}
Given reference images: \texttt{[CONCEPT\_IMAGES]} that all contain the same object, referred to as \textbf{"this subject"}. \\
\textbf{\# TASK DESCRIPTION}\\
\vspace{1pt}
Your task is to generate a detailed description of the object involved in the task.\\
Your response should be in JSON format.\\
Note that your answer should be within 200 words.\\\\ 

\textbf{\# OUTPUT FORMAT}\\
\vspace{1pt}
\{ \\
\quad Description: \{detailed content\} \\
\}
\end{tabular}
\end{sectionbox}
\vspace{-2mm}
\caption{The prompt for generating a general description.}
\label{tab:general_concept_description}
\end{minipage}
\end{table*}

\begin{table*}[t!]
\centering
\begin{minipage}{1.0\textwidth}
\vspace{0mm}    
\centering
\begin{sectionbox}[]{Concept QA Generation}
    \centering
      \footnotesize
    \begin{tabular}{p{0.97\textwidth} c}
\textbf{\# BACKGROUND}\\
\vspace{1pt}
Given reference images: \texttt{[CONCEPT\_IMAGES]} that all contain the same object, referred to as \textbf{"sks"}. \\
Here is a detailed description of the "sks": \texttt{[DESC]} \\\\
\textbf{\# TASK DESCRIPTION}\\
\vspace{1pt}
Your task is to generate a multiple-choice question that references **sks** in the scene image. The question should have two options, one of which is the ground truth.\\
Your question should include \texttt{sks} and present a clear choice between the options. \\\\
\vspace{1pt}
\textbf{\# EXAMPLE}\\
\vspace{1pt}
\begin{itemize}
    \item \textbf{Question:} "Which one is the sks — the one on the left or the one on the right?"
    \item \textbf{Options:}
    \begin{itemize}
        \item A: "The one on the left"
        \item B: "The one on the right"
    \end{itemize}
    \item \textbf{Ground truth:} "B"
\end{itemize}
\textbf{\# EXPLANATION}\\
\vspace{1pt}
For the example above, there is another object besides sks in the scene image. By asking which one is the sks, we force the model to identify the object. The model should answer "B" because it correctly identifies the sks.\\
\vspace{1pt}
\textbf{\# OUTPUT FORMAT}\\
\vspace{1pt}
\{ \\
\quad "question": "detailed content", \\
\quad "options": { "A": "option A content", "B": "option B content" }, \\
\quad "gt": "ground truth" \\
\} 
\end{tabular}
\end{sectionbox}
\vspace{-2mm}
\caption{The prompt for generating a multiple-choice question for concept QA question.}
\label{tab:ablation_qa_generation}
\end{minipage}
\end{table*}
\begin{table*}[t!]
\centering
\begin{minipage}{1.0\textwidth}
\vspace{0mm}    
\centering
\begin{sectionbox}[]{Other Object Description Generation}
    \centering
      \footnotesize
    \begin{tabular}{p{0.97\textwidth} c}
\textbf{\# BACKGROUND}\\
\vspace{1pt}
Given reference images: \texttt{[CONCEPT\_IMAGES]} that all contain the same object, referred to as \textbf{"sks"}. \\
Here is a detailed description about the "sks": \texttt{[DESC]}. \\
Given a scene image: \texttt{[SCENE\_IMAGE]}, which contains the \texttt{"sks"} and several surrounding \texttt{"other objects"}. \\\\
\textbf{\# TASK DESCRIPTION}\\
\vspace{1pt}
Your task is to generate a detailed description of the \texttt{"other objects"} involved in the scene. \\
The description must include the relative position of each object with respect to the \texttt{"sks"}, as well as their basic attributes (e.g., shape, size, color, or category). \\
Your answer should be written in natural language and limited to a maximum of 300 words. \\\\

\textbf{\# OUTPUT FORMAT}\\
\vspace{1pt}
\{ \\
\quad "description": \{detailed content\} \\
\}
\end{tabular}
\end{sectionbox}
\vspace{-2mm}
\caption{Prompt for generating detailed descriptions of objects surrounding the concept target.}
\label{tab:ablation_other_object_description}
\end{minipage}
\end{table*}

\begin{table*}[t!]
\centering
\begin{minipage}{1.0\textwidth}
\vspace{0mm}    
\centering
\begin{sectionbox}[]{Confusion Question Generation}
    \centering
    \footnotesize
    \begin{tabular}{p{0.97\textwidth}}
\textbf{\# BACKGROUND} \\
You are an assessment designer. Your job is to create \textbf{misleading} multiple-choice questions that test whether a vision-language model can truly locate the target object \texttt{"sks"} in a scene image, rather than guessing by shortcut. \\\\
\textbf{Resources} \\
\texttt{[CONCEPT\_IMAGES]} – several reference images that all contain the object \texttt{sks}. \\
\texttt{[DESC]} – a detailed textual description of what \texttt{sks} looks like. \\
\texttt{[SCENE IMAGE]} – a new scene image containing \texttt{sks} plus at least one distractor object. \\
\texttt{[OTHER DESC]} – description of what the other objects in the scene look like. \\\\

\textbf{\# TASK REQUIREMENTS} \\
You should try to mislead the responder by giving questions that \textbf{cannot be answered correctly} unless the model \textbf{actually identifies the sks}. All options should be wrong — only \texttt{None of the above} is correct. \\\\

\textbf{\# EXAMPLES (for style)} \\
\vspace{2pt}
\textbf{Example 1 – False attribute} \\
\texttt{"What is the main color of the object in front of the SKS?"} — misleading because nothing is in front of sks.\\
\textbf{Example 2 – Conflicting location} \\
\texttt{"On which side of the sks can you find the object containing red?"} — the SKS itself is red. \\
\textbf{Example 3 – Category confusion} \\
\texttt{"What kind of kitchen item is the SKS?"} — sks is not a kitchen item.\\\\

\textbf{\# OUTPUT FORMAT} \\
Return \textbf{valid JSON} exactly like this: \\
\begin{verbatim}
[
  {
    "question": "...",
    "options": { "A": "...", "B": "...", "C": "...", "D": "None of above" },
    "gt": "D"
  },
  {
    "question": "...",
    "options": { "A": "...", "B": "...", "C": "...", "D": "None of above" },
    "gt": "D"
  }
]
\end{verbatim}

    \end{tabular}
\end{sectionbox}
\vspace{-2mm}
\caption{Prompt for generating confusion QA.}
\label{tab:ablation_confusion_qa}
\end{minipage}
\end{table*}

\section{AI Assistants}
ChatGPT\footnote{https://chat.openai.com/} was used purely with the language of the paper during the writing process, including spell-checking and paraphrasing the authors' original content, without suggesting new content.
Any content generated with the assistant underwent meticulous manual review and subsequently received final approval from the authors.

%% file: tables/appendix/prominence_check.tex
\begin{table}[ht]
\centering
\resizebox{0.48\textwidth}{!}{%
\begin{tabular}{l|cc}
\hline
\textbf{Setting} & \textbf{Ident. Acc. (\%)} & \textbf{Recall} \\
\hline
\textit{Easy → Middle} & 75.21 & 34.57\\
\textit{Hard → Middle} & 12.64  & 14.32 \\
\hline
\end{tabular}
}
\caption{\textbf{Impact of Concept Image Prominence on Downstream Tasks.} We observe using less prominent (Hard) concept images significantly reduces performance.}
\label{tab:prominence_check}
\end{table}

%% file: tables/appendix/confusion_qa_valid.tex
\begin{table}[ht!]
\centering
\resizebox{0.48\textwidth}{!}{%
\begin{tabular}{lcc}
\hline
\textbf{VQA Type} & \textbf{w/ Concept (↓)} & \textbf{w/o Concept (↓)} \\
\hline
Simple VQA & 95.11 & 80.00 \\
Confusion VQA & \textbf{57.66} & \textbf{20.18} \\
\hline
\end{tabular}
}
\caption{\textbf{VQA Quality Comparison} "w/ Concept" refers to the scenario where concept images are provided, the standard setting. "w/o Concept" refers to when concept images are absent, and the model must answer without perceiving the real concept.}
\label{tab:confusion_qa_valid}
\end{table}

%% file: tables/appendix/zero_shot_vlms.tex
\begin{table*}[ht]
\centering
\scalebox{1.0}{  
\begin{tabular}{l|cc|ll|ll|ll}
\toprule
\multicolumn{1}{c}{\textbf{Task Type}} & \multicolumn{2}{c}{{\textbf{Identification}}} & \multicolumn{2}{c}{{\textbf{QA}}} & \multicolumn{4}{c}{\textbf{Caption Generation}} \\
\cmidrule(lr){1-1}
\cmidrule(lr){2-3}
\cmidrule(lr){4-5}
\cmidrule(lr){6-9}

\multicolumn{1}{c}{} & \multicolumn{2}{c}{{\textbf{Accuracy}}} & \multicolumn{2}{c}{{\textbf{Accuracy}}} & \multicolumn{2}{c}{\textbf{Recall}} & \multicolumn{2}{c}{ \textbf{Txt Sim.}} \\

\multicolumn{1}{c}{\textbf{Q. Setting}} & \multicolumn{1}{c}{Single} & \multicolumn{1}{c}{Multi} & \multicolumn{1}{c}{Single} & \multicolumn{1}{c}{Multi} & \multicolumn{1}{c}{Single} & \multicolumn{1}{c}{Multi} & \multicolumn{1}{c}{Single} & \multicolumn{1}{c}{Multi} \\

\midrule
LLAVA \citep{liu2023visual}& 28.1 & 37.9 & 36.8 & 47.9 & 33.2 & 43.6 & 31.6 & 43.7 \\
Qwen 2.5-7B \citep{bai2025qwen2}& 37.25 & 34.50 & 69.09 & 48.56 & 22.31 & 22.86 & 46.06 & 34.95 \\
Llama 3.2-V \citep{llama3d2v}& 42.71 & 28.57 & 69.52 & 50.91 & 15.21 & 16.28 & 42.29 & 33.21 \\
\textbf{InternVL2.5 7B} \citep{chen2024internvl} & \textbf{57.86} & \textbf{53.56} & \textbf{89.36} & \textbf{83.44} & 17.38 & 20.91 & \textbf{56.01} & \textbf{55.83} \\
\bottomrule
\end{tabular}
}
\caption{Performance of different VLMs on OP-Eval.}
\label{tab:ablation_zero_shot}

\end{table*}

%% file: tables/appendix/ablation_computing_cost.tex
\begin{table}[t]
\centering
\resizebox{0.48\textwidth}{!}{%
\begin{tabular}{l|c}
\hline
\textbf{Model Variant} & \textbf{Computing Cost (FLOPs $\times 10^{22}$)} \\
\hline
Omni Only     & 3.56 \\
LoRA Only                & 0.34 \\
Online-PVLM (Ours)        & 5.07 \\
\hline
\end{tabular}
}
\caption{\textbf{Computational Cost Comparison.}}
\label{tab:ablation_computing_costs}
\end{table}

%% file: custom.bib
@article{nguyen2024yo,
  title={Yo'LLaVA: Your Personalized Language and Vision Assistant},
  author={Nguyen, Thao and Liu, Haotian and Li, Yuheng and Cai, Mu and Ojha, Utkarsh and Lee, Yong Jae},
  journal={arXiv preprint arXiv:2406.09400},
  year={2024}
}

@inproceedings{alaluf2024myvlm,
  title={Myvlm: Personalizing vlms for user-specific queries},
  author={Alaluf, Yuval and Richardson, Elad and Tulyakov, Sergey and Aberman, Kfir and Cohen-Or, Daniel},
  booktitle={ECCV},
  pages={73--91},
  year={2024},
  organization={Springer}
}

@article{pi2024personalized,
  title={Personalized visual instruction tuning},
  author={Pi, Renjie and Zhang, Jianshu and Han, Tianyang and Zhang, Jipeng and Pan, Rui and Zhang, Tong},
  journal={arXiv preprint arXiv:2410.07113},
  year={2024}
}

@article{an2024mc,
  title={Mc-llava: Multi-concept personalized vision-language model},
  author={An, Ruichuan and Yang, Sihan and Lu, Ming and Zhang, Renrui and Zeng, Kai and Luo, Yulin and Cao, Jiajun and Liang, Hao and Chen, Ying and She, Qi and others},
  journal={arXiv preprint arXiv:2411.11706},
  year={2024}
}

@article{liu2023visual,
  title={Visual instruction tuning},
  author={Liu, Haotian and Li, Chunyuan and Wu, Qingyang and Lee, Yong Jae},
  journal={NeurIPS},
  volume={36},
  pages={34892--34916},
  year={2023}
}

@article{achiam2023gpt,
  title={Gpt-4 technical report},
  author={Achiam, Josh and Adler, Steven and Agarwal, Sandhini and Ahmad, Lama and Akkaya, Ilge and Aleman, Florencia Leoni and Almeida, Diogo and Altenschmidt, Janko and Altman, Sam and Anadkat, Shyamal and others},
  journal={arXiv preprint arXiv:2303.08774},
  year={2023}
}

@article{liu2024deepseek,
  title={Deepseek-v3 technical report},
  author={Liu, Aixin and Feng, Bei and Xue, Bing and Wang, Bingxuan and Wu, Bochao and Lu, Chengda and Zhao, Chenggang and Deng, Chengqi and Zhang, Chenyu and Ruan, Chong and others},
  journal={arXiv preprint arXiv:2412.19437},
  year={2024}
}

@inproceedings{chen2024internvl,
  title={Internvl: Scaling up vision foundation models and aligning for generic visual-linguistic tasks},
  author={Chen, Zhe and Wu, Jiannan and Wang, Wenhai and Su, Weijie and Chen, Guo and Xing, Sen and Zhong, Muyan and Zhang, Qinglong and Zhu, Xizhou and Lu, Lewei and others},
  booktitle={CVPR},
  pages={24185--24198},
  year={2024}
}

@inproceedings{hudson2019gqa,
  title={Gqa: A new dataset for real-world visual reasoning and compositional question answering},
  author={Hudson, Drew A and Manning, Christopher D},
  booktitle={CVPR},
  pages={6700--6709},
  year={2019}
}

@article{bai2025qwen2,
  title={Qwen2. 5-vl technical report},
  author={Bai, Shuai and Chen, Keqin and Liu, Xuejing and Wang, Jialin and Ge, Wenbin and Song, Sibo and Dang, Kai and Wang, Peng and Wang, Shijie and Tang, Jun and others},
  journal={arXiv preprint arXiv:2502.13923},
  year={2025}
}

@article{li2023llava,
  title={Llava-med: Training a large language-and-vision assistant for biomedicine in one day},
  author={Li, Chunyuan and Wong, Cliff and Zhang, Sheng and Usuyama, Naoto and Liu, Haotian and Yang, Jianwei and Naumann, Tristan and Poon, Hoifung and Gao, Jianfeng},
  journal={NeurIPS},
  volume={36},
  pages={28541--28564},
  year={2023}
}

@inproceedings{li2024truthreader,
  title={TruthReader: Towards Trustworthy Document Assistant Chatbot with Reliable Attribution},
  author={Li, Dongfang and Hu, Xinshuo and Sun, Zetian and Hu, Baotian and Ye, Shaolin and Shan, Zifei and Chen, Qian and Zhang, Min},
  booktitle={EMNLP},
  pages={89--100},
  year={2024}
}

@article{yunusov2024mirrorstories,
  title={MirrorStories: Reflecting Diversity through Personalized Narrative Generation with Large Language Models},
  author={Yunusov, Sarfaroz and Sidat, Hamza and Emami, Ali},
  journal={arXiv preprint arXiv:2409.13935},
  year={2024}
}

@article{hu2022lora,
  title={Lora: Low-rank adaptation of large language models.},
  author={Hu, Edward J and Shen, Yelong and Wallis, Phillip and Allen-Zhu, Zeyuan and Li, Yuanzhi and Wang, Shean and Wang, Lu and Chen, Weizhu and others},
  journal={ICLR},
  volume={1},
  number={2},
  pages={3},
  year={2022}
}

@inproceedings{hong2024cogagent,
  title={Cogagent: A visual language model for gui agents},
  author={Hong, Wenyi and Wang, Weihan and Lv, Qingsong and Xu, Jiazheng and Yu, Wenmeng and Ji, Junhui and Wang, Yan and Wang, Zihan and Dong, Yuxiao and Ding, Ming and others},
  booktitle={CVPR},
  pages={14281--14290},
  year={2024}
}

@article{yang2025r1,
  title={R1-onevision: Advancing generalized multimodal reasoning through cross-modal formalization},
  author={Yang, Yi and He, Xiaoxuan and Pan, Hongkun and Jiang, Xiyan and Deng, Yan and Yang, Xingtao and Lu, Haoyu and Yin, Dacheng and Rao, Fengyun and Zhu, Minfeng and others},
  journal={arXiv preprint arXiv:2503.10615},
  year={2025}
}

@inproceedings{plummer2015flickr30k,
  title={Flickr30k entities: Collecting region-to-phrase correspondences for richer image-to-sentence models},
  author={Plummer, Bryan A and Wang, Liwei and Cervantes, Chris M and Caicedo, Juan C and Hockenmaier, Julia and Lazebnik, Svetlana},
  booktitle={ICCV},
  pages={2641--2649},
  year={2015}
}

@article{zhang2021prototypical,
    title={Prototypical Pseudo Label Denoising and Target Structure Learning for Domain Adaptive Semantic Segmentation},
    author={Zhang, Pan and Zhang, Bo and Zhang, Ting and Chen, Dong and Wang, Yong and Wen, Fang},
    journal={arXiv preprint arXiv:2101.10979},
    year={2021}
}

@inproceedings{zhou2022cocoop,
    title={Conditional Prompt Learning for Vision-Language Models},
    author={Zhou, Kaiyang and Yang, Jingkang and Loy, Chen Change and Liu, Ziwei},
    booktitle={CVPR},
    year={2022}
}

@inproceedings{Wang2024LearningTL,
  title={Learning to Learn Better Visual Prompts},
  author={Fengxiang Wang and Wanrong Huang and Shaowu Yang and Qi Fan and Long Lan},
  booktitle={AAAI},
  year={2024},
  url={https://api.semanticscholar.org/CorpusID:268692579}
}

@article{zhang2021tip,
  title={Tip-Adapter: Training-free CLIP-Adapter for Better Vision-Language Modeling},
  author={Zhang, Renrui and Fang, Rongyao and Gao, Peng and Zhang, Wei and Li, Kunchang and Dai, Jifeng and Qiao, Yu and Li, Hongsheng},
  journal={arXiv preprint arXiv:2111.03930},
  year={2021}
}

@article{gao2021clip,
  title={CLIP-Adapter: Better Vision-Language Models with Feature Adapters},
  author={Gao, Peng and Geng, Shijie and Zhang, Renrui and Ma, Teli and Fang, Rongyao and Zhang, Yongfeng and Li, Hongsheng and Qiao, Yu},
  journal={arXiv preprint arXiv:2110.04544},
  year={2021}
}

@misc{song2019combining,
      title={Combining MixMatch and Active Learning for Better Accuracy with Fewer Labels},
      author={Shuang Song and David Berthelot and Afshin Rostamizadeh},
      year={2019},
      eprint={1912.00594},
      archivePrefix={arXiv},
      primaryClass={cs.LG}
}

@misc{hao2024rememberretrievegenerateunderstanding,
        title={Remember, Retrieve and Generate: Understanding Infinite Visual Concepts as Your Personalized Assistant}, 
        author={Haoran Hao and Jiaming Han and Changsheng Li and Yu-Feng Li and Xiangyu Yue},
        year={2024},
        eprint={2410.13360},
        archivePrefix={arXiv},
        primaryClass={cs.CV},
        url={https://arxiv.org/abs/2410.13360}, 
  }

@article{qian2024scaling,
  title={Scaling large-language-model-based multi-agent collaboration},
  author={Qian, Chen and Xie, Zihao and Wang, Yifei and Liu, Wei and Dang, Yufan and Du, Zhuoyun and Chen, Weize and Yang, Cheng and Liu, Zhiyuan and Sun, Maosong},
  journal={arXiv preprint arXiv:2406.07155},
  year={2024}
}

@misc{kingma2017adammethodstochasticoptimization,
      title={Adam: A Method for Stochastic Optimization}, 
      author={Diederik P. Kingma and Jimmy Ba},
      year={2017},
      eprint={1412.6980},
      archivePrefix={arXiv},
      primaryClass={cs.LG},
      url={https://arxiv.org/abs/1412.6980}, 
}

@misc{gallegos2024biasfairnesslargelanguage,
      title={Bias and Fairness in Large Language Models: A Survey}, 
      author={Isabel O. Gallegos and Ryan A. Rossi and Joe Barrow and Md Mehrab Tanjim and Sungchul Kim and Franck Dernoncourt and Tong Yu and Ruiyi Zhang and Nesreen K. Ahmed},
      year={2024},
      eprint={2309.00770},
      archivePrefix={arXiv},
      primaryClass={cs.CL},
      url={https://arxiv.org/abs/2309.00770}, 
}

@misc{guo2024biaslargelanguagemodels,
      title={Bias in Large Language Models: Origin, Evaluation, and Mitigation}, 
      author={Yufei Guo and Muzhe Guo and Juntao Su and Zhou Yang and Mengqiu Zhu and Hongfei Li and Mengyang Qiu and Shuo Shuo Liu},
      year={2024},
      eprint={2411.10915},
      archivePrefix={arXiv},
      primaryClass={cs.CL},
      url={https://arxiv.org/abs/2411.10915}, 
}

@misc{mettes2023hyperbolicdeeplearningcomputer,
      title={Hyperbolic Deep Learning in Computer Vision: A Survey}, 
      author={Pascal Mettes and Mina Ghadimi Atigh and Martin Keller-Ressel and Jeffrey Gu and Serena Yeung},
      year={2023},
      eprint={2305.06611},
      archivePrefix={arXiv},
      primaryClass={cs.CV},
      url={https://arxiv.org/abs/2305.06611}, 
}

@inproceedings{li-etal-2022-hypoformer,
    title = "Hypoformer: Hybrid Decomposition Transformer for Edge-friendly Neural Machine Translation",
    author = "Li, Sunzhu  and
      Zhang, Peng  and
      Gan, Guobing  and
      Lv, Xiuqing  and
      Wang, Benyou  and
      Wei, Junqiu  and
      Jiang, Xin",
    editor = "Goldberg, Yoav  and
      Kozareva, Zornitsa  and
      Zhang, Yue",
    booktitle = "NIPS",
    month = dec,
    year = "2022",
    address = "Abu Dhabi, United Arab Emirates",
    publisher = "Association for Computational Linguistics",
    url = "https://aclanthology.org/2022.emnlp-main.475/",
    doi = "10.18653/v1/2022.emnlp-main.475",
    pages = "7056--7068"
}

@article{DBLP:journals/corr/NickelK17,
  author       = {Maximilian Nickel and
                  Douwe Kiela},
  title        = {Poincar{\'{e}} Embeddings for Learning Hierarchical Representations},
  journal      = {CoRR},
  volume       = {abs/1705.08039},
  year         = {2017},
  url          = {http://arxiv.org/abs/1705.08039},
  eprinttype    = {arXiv},
  eprint       = {1705.08039},
  bibsource    = {dblp computer science bibliography, https://dblp.org}
}

@misc{reimers2019sentencebertsentenceembeddingsusing,
      title={Sentence-BERT: Sentence Embeddings using Siamese BERT-Networks}, 
      author={Nils Reimers and Iryna Gurevych},
      year={2019},
      eprint={1908.10084},
      archivePrefix={arXiv},
      primaryClass={cs.CL},
      url={https://arxiv.org/abs/1908.10084}, 
}

@misc{llama3d2v,
    title={Llama 3.2: Revolutionizing edge AI and vision with open, customizable models},
    url={https://ai.meta.com/blog/llama-3-2-connect-2024-vision-edge-mobile-devices/},
    month={Sep},
    year={2024},
    author={Google}
}

@misc{openai2024o1,
  author       = {OpenAI},
  title        = {Introducing OpenAI o1},
  year         = {2024},
  howpublished = {\url{https://openai.com/o1/}},
  note         = {Accessed: 2025-05-16}
}

@INPROCEEDINGS{4109404,
  author={Hamaguchi, Kiyoshi and Ogawa, Hiroyo},
  booktitle={IEEE Vehicular Technology Conference}, 
  title={Design and Performance of 32 GHz Portable Broadband Wireless Access System}, 
  year={2006},
  volume={},
  number={},
  pages={1-5},
  doi={10.1109/VTCF.2006.139}}
